\pdfoutput=1

\documentclass[11pt]{article}

\usepackage[preprint]{acl}

\usepackage{times}
\usepackage{latexsym}

\usepackage[T1]{fontenc}

\usepackage[utf8]{inputenc}

\usepackage{microtype}

\usepackage{inconsolata}

\usepackage{tabularx} 
\usepackage{listings}
\usepackage{threeparttable}

\usepackage{times}
\usepackage{soul}
\usepackage{url}

\usepackage{booktabs}
\usepackage{tabularx}
\usepackage{multirow}
\usepackage{array}
\usepackage{placeins}
\usepackage{listings}

\usepackage{graphicx}
\setkeys{Gin}{draft=false} 
\usepackage{amsmath}
\usepackage{amsthm}
\usepackage{booktabs}
\usepackage{algorithm}
\usepackage{algorithmic}
\usepackage{inconsolata}
\usepackage{import}
\usepackage{microtype}
\usepackage{layout}
\usepackage{tabularx, makecell}
\usepackage{booktabs}
\usepackage{mathrsfs}
\usepackage{amssymb} 
\usepackage{url}
\usepackage{graphicx}
\usepackage{bbm}
\usepackage{xspace,paralist}
\usepackage{times,latexsym}
\usepackage{amsmath}
\usepackage{appendix}
\usepackage{comment} 
\usepackage{enumitem}
\usepackage{makecell}
\usepackage{multirow}
\usepackage{xcolor}
\usepackage{graphicx}
\usepackage{cleveref}

\usepackage{adjustbox}
\usepackage{relsize}
\usepackage{todonotes}
\usepackage{natbib}
\usepackage{CJKutf8}
\usepackage{colortbl}
\usepackage{subcaption}
\usepackage{booktabs}
\usepackage{adjustbox}
\usepackage{booktabs}
\usepackage{multirow}
\usepackage{xcolor}
\usepackage{colortbl}
\usepackage{float}
\usepackage[most]{tcolorbox}

\tcbuselibrary{breakable,listings,skins}

\definecolor{headerblue}{RGB}{30,64,124}
\definecolor{headergray}{RGB}{235,238,245}
\definecolor{ourrow}{RGB}{254,243,199}
\definecolor{subtlegray}{RGB}{248,249,250}
\definecolor{sectionbg}{RGB}{235,238,245}
\definecolor{sectionbg}{RGB}{235,238,245}
\definecolor{ourrow}{RGB}{254,243,199}
\definecolor{subtlegray}{RGB}{248,249,250}
\definecolor{baselinerow}{RGB}{237,242,251}
\definecolor{lightyellow}{RGB}{255,249,230} 
\definecolor{sectionbg}{RGB}{235,238,245}
\definecolor{ourrow}{RGB}{254,243,199}

\newcommand{\secref}[1]{Section~\ref{#1}\xspace}

\usepackage{fvextra}

\DefineVerbatimEnvironment{PromptBlock}{Verbatim}{%
  fontsize=\scriptsize,
  frame=single,
  framesep=2.5mm,
  framerule=0.3pt,
  breaklines=true,
  breakanywhere=true,
  baselinestretch=0.95,
  xleftmargin=0pt,
  xrightmargin=0pt
}

\usepackage{pifont}
\usepackage{listings}
\usepackage{tcolorbox}
\usepackage{colortbl}
\usepackage[table]{xcolor}
\definecolor{yellow}{RGB}{255, 255, 150}      
\definecolor{lightblue}{RGB}{173, 216, 230}   
\definecolor{lightred}{RGB}{255, 182, 193}    
\definecolor{lightgreen}{RGB}{144, 238, 144}  

\definecolor{greenbox}{RGB}{144, 238, 144}    
\definecolor{redbox}{RGB}{255, 182, 193}      
\definecolor{bluebox}{RGB}{135, 206, 235}     
\definecolor{yellowbox}{RGB}{255, 255, 0}     
\definecolor{posgreen}{RGB}{198, 239, 206}  
\definecolor{negred}{RGB}{255, 199, 206}    

\usepackage[table]{xcolor}
\usepackage{array}

\definecolor{bestO}{HTML}{E8F0E5}    
\definecolor{bestR}{HTML}{DCE6F2}    
\definecolor{rowshade}{HTML}{F7F7F7} 

\newcommand{\cbO}[1]{\cellcolor{bestO}#1}      
\newcommand{\cbR}[1]{\cellcolor{bestR}#1}      

\usepackage{threeparttable}
\usepackage{multicol}
\usepackage{graphicx}
\usepackage{tcolorbox}
\usepackage{xcolor}
\usepackage{CJKutf8} 
\usepackage[utf8]{inputenc}

\definecolor{greenbox}{HTML}{D9EAD3}  
\definecolor{bluebox}{HTML}{D9EAF7}   
\definecolor{redbox}{HTML}{F4CCCC}    
\definecolor{yellowbox}{HTML}{FFF2CC} 

\usepackage[ruled,linesnumbered,algo2e]{algorithm2e}

\tcbset{
    eval_prompt/.style={
        colback=gray!5,
        colframe=black!70,
        boxrule=0.5pt,
        arc=2mm,
        left=8pt, right=8pt, top=8pt, bottom=8pt,
        fontupper=\small,
        width=\linewidth
    }
}

\title{AI YOU Town: Make Friends and Money with Your Digital Twin}

\author{
\textbf{Yan Lin}\textsuperscript{1,2}\thanks{\ \ Equal contribution.},
\textbf{Yuyang Dai}\textsuperscript{1}\footnotemark[1],
\textbf{Jiahui Geng}\textsuperscript{3},
\textbf{Yuxia Wang}\textsuperscript{1} \\ 
\textsuperscript{1}INSAIT, Sofia University ``St. Kliment Ohridski'' \\
\textsuperscript{2}Newcastle University \quad
\textsuperscript{3}Link\"oping University \\
\texttt{y.lin64@ncl.ac.uk, y9657422@gmail.com,} \\
\texttt{jiahui.geng@liu.se, yuxia.wang@insait.ai}
}

\begin{document}
\maketitle
\begin{abstract}
Existing approaches to infer user traits and generate responses consistent with a persona rely on static prompting. They lack calibrated uncertainty, ignore sequential evidence, and drift during long interactions. We present \textbf{AI YOU}, a framework that continually updates a personality profile with 22 dimensions from conversation and embodies it in a personal digital twin. 
Practically, the system combines prompting, Bayesian updating, and conformal prediction for persona inference. A periodically refreshed memory anchor and cognitive memory with three layers preserve persona consistency over long interactions.
Across the main results, AI YOU \emph{(i)} achieves conformal coverage ranging from 0.921 to 0.976, \emph{(ii)} improves uncertainty calibration and reasoning grounded in memory, and \emph{(iii)} enhances persona fidelity over static prompting in role playing over 100 turns while reducing trait drift, for most evaluated backbones under adversarial settings with multiple agents.
The prototype \emph{AI YOU Town} initializes an imaginative twin world for future interaction. The online demo is available at \href{https://quinnnnnne-ai-you.hf.space/}{\mbox{\texttt{quinnnnnne-ai-you.hf.space}}}.

\end{abstract}

\section{Introduction}

Imagine a virtual town populated by personal digital twins that interact, socialize, and work on behalf of their users while preserving users' personalities.
Large Language Models (LLMs) have shown promising ability to infer personality traits from text and generate dialogue that reflects individual communicative styles \cite{shao2023character, park2023generative, peters2024large, jiang2024personallm, marengo2025inferring}, making such systems increasingly plausible.
However, building a reliable personal digital twin requires solving three intertwined challenges that existing systems address only in isolation.

\begin{figure*}[t]
    \centering
    \includegraphics[width=\textwidth]{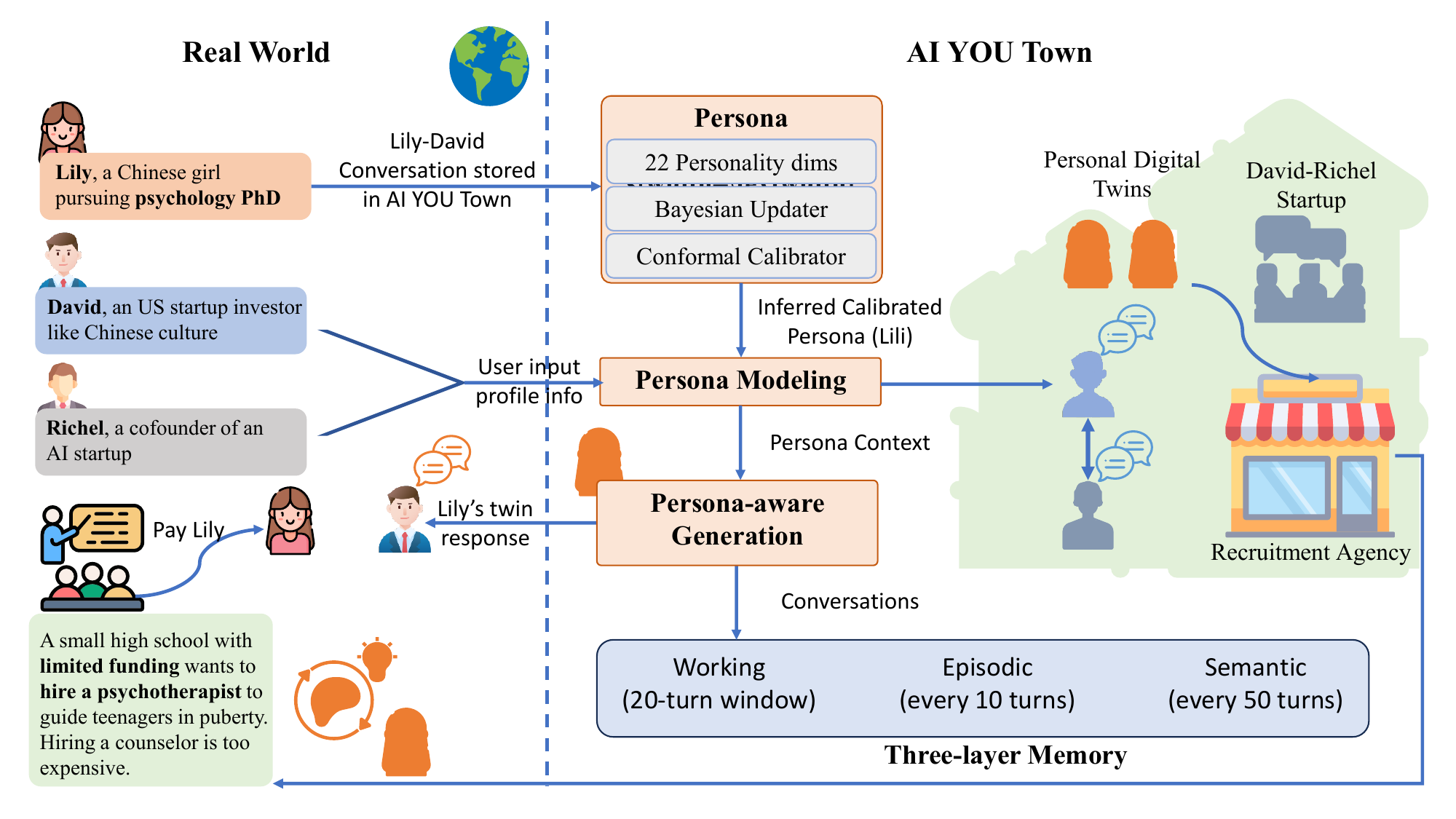}
    \caption{\textbf{AI YOU} Overview: persona inference, modeling, persona-aware generation based on three-layer memory.}
    \label{fig:overview}
\end{figure*}

\textit{First,} personality inference from conversation requires \textit{uncertainty-aware} estimation: single-pass LLM predictions produce point estimates without calibrated confidence \cite{song2023have, lin2025prompts}.
\textit{Second,} personality evidence accumulates across turns, yet most pipelines treat utterances independently, lacking mechanisms for sequential belief updating \cite{tan2025prospect, chen2025deeper}.
\textit{Third,} personality inference and persona-driven generation are typically treated as separate problems \cite{tseng2024two, zhong2025evaluating}, creating a gap between \textit{understanding} and \textit{embodying} a user.

Therefore, we present \textbf{AI YOU}, an interactive multi-agent framework that organically unifies isolated modules in one system. Modules interact cohesively: personality inference outputs serve as the input of persona modeling, persona-conditioned generations support persona updating, and a shared three-layer memory enables persona consistency and long-horizon reasoning. It is related to LLM-based personality simulation, dynamic persona inference, persona-conditioned generation, and digital-twin simulation. We provide a detailed related work in Section~\ref{sec:related}.

We further prototype the \textbf{AI YOU Town}, a virtual environment where personal digital twins (PDTs) interact, socialize, and work.
As illustrated in Figure~\ref{fig:overview}, it is a bidirectional ecosystem that connects the real world and AI YOU Town. 
The story begins when a real user initiates a conversation with either a twin or another real user logged into the town.
Then, a specialized agent continually estimates a 22-dimensional profile based on user conversations, where Bayesian posterior inference and conformal prediction maintain calibrated uncertainty. The inferred persona would be used to initialize or stabilize the user's digital twin, and a persona-conditioned agent that reproduces the inferred behavioral patterns in new interactions.
We design a three-layer memory architecture (working, episodic, and semantic) that grounds responses in accumulated interaction history, to support persona consistency and long-horizon reasoning. 
To summarize, our contributions are as below:

\begin{compactitem}
    \item We design the initial prototype of a personal digital twin town: AI YOU, which organically connects the real world and enables two functions: companion and consulting, leaving trading as an imaginative hook. 
    \item We unify personality inference, persona modeling and preservation, and persona-conditioned generation within a single system, where a grounded three-layer memory mechanism enables persona consistency and long-horizon reasoning.
    \item Experiments show that the methods used in each module achieve superior performance over strong baselines, demonstrating high accuracy, and uncertainty estimation via Bayesian posterior inference and conformal prediction further enhances reliability.
\end{compactitem}
\section{Related Work}
\label{sec:related}

\subsection{LLM-Based Personality Simulation}
\label{app:rw-simulation}

Large language models are increasingly used to simulate human-like personalities, enabling more engaging social agents~\cite{chen2024persona}. Three questions naturally arise:
\emph{(i)} What traits should be included in personality modeling?
\emph{(ii)} How can an LLM be conditioned on a specific persona?
\emph{(iii)} How can persona consistency over time be evaluated, and through which tasks or applications can we rigorously assess its persistence?

\vspace{-0.2cm}
\paragraph{Persona Traits}
\citet{jiang2024personallm} applied the Big Five traits to model and evaluate simulated personality,\footnote{Big Five is a model of personality that describes people based on five key traits: openness, conscientiousness, extraversion, agreeableness, and neuroticism.} which remains the dominant framework for LLM-based personality modeling~\cite{serapio2023personality, wang2025evaluating}. Recent work adopts psychometric validation protocols using standardized assessments such as IPIP-NEO~\cite{song2023have}, and proposes construct-grounded methods for more principled personality design~\cite{huang2026designing, lin2025prompts}. However, the Big Five alone cannot fully capture individual differences. It omits interpersonal constructs (e.g.,\ attachment style), wellbeing-related aspects (e.g.,\ self-efficacy, perceived loneliness), and emotional states (e.g.,\ positive and negative affect). Recent work further shows that LLMs can model psychological structures beyond the Big Five, predicting responses on nine additional validated scales from minimal personality inputs~\cite{liu2025fivetomany}. In our work, we therefore extend beyond the Big Five to a 22-dimensional psychological profile. 

\vspace{-0.2cm}
\paragraph{Enabling and Preserving Persona}
Early work on dialogue personalization relied on meta-learning~\cite{madotto2019personalizing}, while recent generative language models enable zero-shot prompt control and character-level fine-tuning~\cite{ramirez2023controlling, shao2023character, serapio2023personality}. Among these approaches, system prompting has become the dominant method for persona conditioning because it requires no additional training and allows real-time, user-specific instantiation~\cite{jiang2024personallm}. Empirical studies show that prompt-assigned personas are reflected in both personality test scores and generated linguistic patterns~\cite{jiang2024personallm}. However, prompt-assigned personas often drift during extended conversations: models contradict earlier statements, drift in tone, and gradually abandon role-appropriate behavior~\cite{li2024measuring, choi2024identity}. Simply assigning a persona at the beginning is insufficient to prevent this drift, and larger models may exhibit greater identity instability than smaller ones~\cite{choi2024identity}. Recent work proposes multi-turn reinforcement learning to reduce persona inconsistency by over 55\%~\cite{abdulhai2025consistently}, but this requires costly fine-tuning for each persona. Further challenges include limited cross-cultural adaptability~\cite{kwok2024evaluating}. In our work, we address persona preservation through structured long-term memory and a three-layer cognitive architecture, improving persona consistency across extended interactions without per-persona fine-tuning.

\vspace{-0.3cm}
\paragraph{Persona Consistency Evaluation}
Most prior work evaluates persona consistency through static post-hoc assessments such as personality questionnaires or single-pass trait scoring~\cite{song2023have, wang2025evaluating}. However, persona consistency in real interaction is inherently sequential: user beliefs evolve over time as new conversational evidence accumulates. Our work therefore evaluates persona consistency through real-time probing and sequential Bayesian updating, dynamically maintaining trait estimates and uncertainty throughout multi-turn interaction.
Our work combines these three aspects within a unified framework. 
We model personality using a 22-dimensional psychological profile extending beyond the Big Five, instantiate personas through system prompting, and evaluate persona consistency through sequential Bayesian updating and real-time probing across conversation turns~\cite{jiang2024personallm, serapio2023personality, song2023have, wang2025evaluating}.



\vspace{-0.3cm}
\subsection{Personality Inference}
\label{app:rw-inference}
Personality inference needs multiple turns, but most LLM systems treat each utterance independently and lack a persistent user model.
Recent work addresses this along three directions. 
First, \textit{heuristic reflection} methods summarize and refine user representations over dialogue history \cite{tan2025prospect}, using multi-granularity memory and iterative persona refinement \cite{chen2025deeper}. 
Second, \textit{structured memory architectures} organize evidence into multiple stores \cite{li2025hello}, including event-level, episodic, and persona representations \cite{yan2026adamem}, and have been applied to long-horizon interaction and multi-agent systems \cite{westhausser2025enabling}. 
Third, \textit{probabilistic approaches} model belief updating explicitly, either by analyzing deviations from Bayesian reasoning \cite{qiu2026bayesian} or by maintaining external posteriors for preference inference \cite{austin2024bayesian}.

\vspace{-0.2cm}
\paragraph{Uncertainty Quantification for LLM Predictions}
A key limitation of single-pass LLM inference is the lack of calibrated uncertainty: model confidence scores are often miscalibrated and lack formal guarantees \cite{kumar2023conformal}. 
Conformal prediction \cite{angelopoulos2021gentle} provides a model-agnostic, distribution-free framework that converts heuristic uncertainty into prediction sets with coverage guarantees. 
Recent work extends conformal methods to NLP settings, including API-only LLMs and open-ended generation \cite{campos2024conformal, su2024api, wang2025conu}. 
AI YOU addresses this gap by combining Bayesian updating with Adaptive Prediction Sets to produce per-dimension prediction sets under sequential evidence accumulation.

\vspace{-0.2cm}
\subsection{Persona-Conditioned Generation}
\label{app:rw-generation}


In human-model interaction, persona-conditioned generation typically takes two forms, which \citet{tseng2024two} term \textit{personalization} and \textit{role-playing}.

The first, \textit{personalization}, generates responses aligned with the user's preferences, needs, and communication style~\cite{salemi2024lamp, tan2025prospect}. The primary challenge is accurate user modeling, since generation quality depends on the quality of the inferred personality profile, while user signals are often sparse, heterogeneous, and noisy~\cite{chen2024persona}. Prior work addresses this through retrieval-augmented generation, user-history conditioning, and online preference learning~\cite{li2025hello, chen2025deeper}.

The second, \textit{role-playing}, conditions the model on a fixed persona, requiring consistent behavior aligned with a specified personality rather than a generic assistant style. This is commonly implemented through system prompting, a training-free approach that is simple, flexible, and transferable across models~\cite{tseng2024two, chen2024persona}. Alternatively, supervised fine-tuning and alignment training (e.g., SFT, RLHF) can improve persona fidelity, but at substantial computational cost, limited cross-persona generalizability, and potential degradation of general capabilities~\cite{shao2023character, serapio2023personality}. We adopt system prompting because it supports real-time persona instantiation without additional training, consistent with findings that prompt-based control achieves trait-consistent behavior comparable to fine-tuned alternatives~\cite{jiang2024personallm, ramirez2023controlling}. The main challenge is maintaining persona consistency over long interactions, as off-the-shelf LLMs often drift from assigned personas, contradict earlier statements, or abandon role-appropriate behavior~\cite{li2024measuring, choi2024identity, abdulhai2025consistently}.

Our framework integrates both interpretations within a unified pipeline. It combines continuous user modeling for personalization with long-term memory mechanisms for persona preservation. We further distinguish among three persona types: \textit{demographic archetypes, fictional characters, }and\textit{ real individuals}, each introducing different simulation challenges (\secref{app:rw-persona-types}).

\vspace{-0.2cm}
\subsection{Differences of Three Types of Personas}
\label{app:rw-persona-types}

\paragraph{Demographic personas}
Demographic simulation assigns group-level attributes such as age, gender, occupation, and cultural background to an LLM, and evaluates whether its outputs reflect the expected distributional patterns of that group~\cite{argyle2023silicon, santurkar2023opinions}. The primary challenge is avoiding stereotyping and caricature: LLMs often produce homogeneous and socially desirable outputs that fail to reflect real intra-group diversity, while simulation fidelity varies substantially across demographic axes~\cite{cheng2023marked, peng2025llm}. This type is inherently cross-sectional, capturing population-level tendencies rather than the behavior of a specific individual over time.

\vspace{-0.2cm}
\paragraph{Character personas}
Character simulation targets fictional or public figures, requiring the model to rely on stored knowledge about a person's speech style, background, and behavioral patterns~\cite{shao2023character, chen2024persona}. The main challenge is character hallucination: models may generate plausible but factually inconsistent behavior when knowledge about the character is incomplete~\cite{lu2024large}. Like demographic simulation, character simulation treats persona traits as largely fixed and does not model long-term behavioral evolution.

\vspace{-0.2cm}
\paragraph{Individualized personas}
Simulating a \textit{real individual} is fundamentally different from the above two settings. Demographic and character simulation both rely on externally specified traits, whereas an individualized simulation, which we term a Personal Digital Twin (PDT), must instead model longitudinal behavior, adaptive responses, and evolving preferences~\cite{saracco2019digital, fuchs2023modeling}. This distinction is particularly important in companion applications, where demographic information alone is often insufficient for accurate personalization~\cite{huang2024simulation}. 

Recent work shows that fine-grained individual simulation is feasible: models conditioned on interview transcripts can reproduce individual survey responses at near test-retest reliability~\cite{park2024generative}, while probabilistic digital twin frameworks maintain uncertainty-aware latent representations of user identity~\cite{david2025probabilistic}. Crucially, inference and simulation form an iterative loop: simulated interactions generate new behavioral evidence, which updates the inferred profile and improves future simulation fidelity. This bidirectional process distinguishes PDTs from demographic and character simulation.
We primarily focus on twining individuals.

\vspace{-0.3cm}
\section{AI-YOU Town}
\label{sec:framework}


AI YOU Town is a virtual environment, where users in the real world establish their personal digital twins (PDTs). Twins can communicate with real users as friends and can also be employed by companies in real world in much lower price. As shown in Figure~\ref{fig:overview}, Lily's twin is the friend of David and also serves as a counselor in high school. 

Twins can be initiated by users who proactively fill out persona modeling features (e.g.,\ demographic, psychological), like David and Richel in Figure~\ref{fig:overview}.
Twins also can be passively established based on system-inferred persona according to users' conversation and behavior patterns stored in the memory, like Lily. 
Regardless of the initialization point, all personas will be updated based on the new conversations and actions, evolving over time. Meanwhile, once the persona is accurately modeled, we preserve the persona across conversations and behaviors, simulating humans' personality consistency.

To achieve functions above and safeguards, our system consists of three major components including \emph{(i)} persona inference, \emph{(ii)} persona modeling and preservation, and \emph{(iii)} affective, relationship and risk monitoring, along with a three-layer memory mechanism supporting long-term learning. 

\subsection{Persona Inference}
Persona inference aims to estimate demographic attributes, personality traits, communication style, affective states, and interpersonal preferences from user's dialogue history stored in memory. 
Formally, given the dialogue history $D_{1:t}$ and retrieved memory $M_t$ at time $t$, the estimator produces a structured profile $\hat{z}_t$ covering 22 dimensions: Big Five scores in $[0,1]$; Other fields are set to \texttt{null} when evidence is insufficient. 
The inferred profile serves two downstream roles: \emph{(i)} initialize and continually update the user's personal digital twin (Section~\ref{sec:persona_modeling}), \emph{(ii)} condition persona-aware generation. 
Specifically, we infer the following dimensions.

\paragraph{Persona Profiling Features}
The profiling features three layers. The \textit{first layer} captures demographic attributes (e.g.,\ gender, age, occupation, education, geographic region), which provide contextual grounding for downstream reasoning but are treated as optional and uncertain.
The \textit{second layer} encodes validated psychometric constructs from personality and clinical psychology, including the Big Five \cite{john1999big}, adult attachment (anxiety and avoidance) \cite{hazan2017romantic, brennan1998self}, self-efficacy \cite{bandura1977self, schwarzer1995generalized}, perceived loneliness \cite{russell1996ucla}, and emotional state (positive and negative affect) \cite{watson1988development}.
These constructs are used for quantitative inference, evidence attribution, and uncertainty estimation.
The \textit{third layer} includes popular self-report frameworks commonly used in conversational settings, specifically the Myers--Briggs Type Indicator (MBTI) \cite{mccrae1989reinterpreting}.
We include MBTI to improve user interpretability, because many users are familiar with it but not with the Big Five, while acknowledging its contested psychometric validity \cite{pittenger1993utility, stein2019evaluating}.

\paragraph{Persona Estimation}

Given the heterogeneity of the target schema: some attributes are numeric, some are categorical, some are free-text descriptors, and many are missing or only weakly supported in short conversations, we applied two prompting-based strategies (\textit{Direct}, \textit{CoT}) rather than a task-specific supervised classifier to estimate persona.

\emph{Direct} prompt instructs the model to use \texttt{null} under insufficient evidence, defines confidence $0$ as no evidence and confidence $1$ as strong evidence, and constrains Big Five and interest scores to the $0$--$1$ interval. 
\emph{CoT} variant first asks the model to inspect topics, linguistic style, values, goals, Big Five cues, and interests before emitting the final outputs. We report Direct and CoT results separately in Section~\ref{sec:experiments}. Detailed prompts are provided in Appendix~\ref{app:prompts}.
When sufficient context is available, the estimator additionally performs contrastive extraction: it compares an estimate from the full recent context with an estimate from a minimal recent context and downweights attributes that appear only under weak contextual support. This step reduces spurious profile updates caused by isolated turns.

\paragraph{Uncertainty Estimation}

AI YOU maintains uncertainty through Bayesian posterior estimation and conformal prediction. For sequential numeric observations, we use a Gaussian conjugate update. Given an observation $x_t$ with confidence score $c_t$, the observation variance is defined as
\begin{equation*}
\sigma_{\mathrm{obs}}^2=\max(10^{-3}, 1-c_t).
\end{equation*}
Given prior mean $\mu_{t-1}$ and variance $\sigma_{t-1}^2$, the posterior is updated by precision-weighted averaging:
\begin{equation*}
\sigma_t^{-2}=\sigma_{t-1}^{-2}+\sigma_{\mathrm{obs}}^{-2}, \quad
\mu_t=\sigma_t^2\left(\frac{\mu_{t-1}}{\sigma_{t-1}^2}+\frac{x_t}{\sigma_{\mathrm{obs}}^2}\right).
\end{equation*}
Categorical variables are updated discretely rather than averaged: when conflicting labels are observed, the system retains the label with stronger evidence and confidence.
For set-valued uncertainty, we apply conformal calibration with $\alpha=0.10$. The nonconformity score is defined as $1-p(y_{\mathrm{true}})$, where $p(y_{\mathrm{true}})$ is the probability assigned to the true label. Calibration is grouped by turn bucket and personality dimension, with a global fallback for groups with insufficient calibration data. Age, Big Five dimensions, and interests are discretized into bins for prediction-set construction. The final output includes point predictions, calibrated prediction sets, empirical coverage, average set size, and singleton counts.

\subsection{Affective, Relationship, Risk Monitor}
\label{sec:auxiliary}

Persona inference alone is insufficient for safe long-term interaction. AI YOU therefore maintains auxiliary monitors for affective state tracking, relationship modeling, and risk detection throughout multi-turn conversations. These monitors are essential for maintaining relational health and safety: affective and relationship tracking helps users avoid misunderstandings and make informed relational decisions, while risk monitoring identifies unsafe manipulation, including online fraud, romance scams, and pig-butchering investment schemes. For example, given evidence of user financial stress and a PDT suggesting high-return investments, the monitor predicts an \textit{anxious/high-intensity} affective state $E_t$, an \textit{asymmetric persuasion} relationship state $R_t$, and a \textit{high-risk} state $R^{\mathrm{risk}}_t$.

\textbf{Affective Monitor:}
it estimates the user's current emotional state, emotional intensity, and likely short-term emotional trajectory from the current utterance and recent emotional history. The method combines an LLM-based structured classifier with a history-aware transition model. The resulting affective state $E_t$ conditions downstream response strategies such as validation, low-pressure follow-up, boundary setting, and risk-sensitive support.

\textbf{Relationship Monitor:}
it models interpersonal dynamics from dialogue context, emotional history, and compressed interaction logs. The system tracks relationship sentiment, interaction status, and potential relationship progression while retaining uncertain interpretations as hypotheses rather than stable facts unless supported by repeated evidence. The resulting relationship state $R_t$ is used for long-term interaction planning and personalization.

\textbf{Risk Monitor:}
it detects potentially unsafe interaction patterns from the current turn and conversation history. The method combines rule-based pattern matching with LLM-based semantic analysis to identify manipulation, coercive interaction, romance scams, and unsafe emotional dependency. When the estimated risk state $R^{\mathrm{risk}}_t$ exceeds a threshold, the coordinator can issue warnings, restrict unsafe personalization behaviors, or request additional verification before continuing.

\subsection{Persona Modeling and Preservation}
\label{sec:persona_modeling}

Given a calibrated persona profile $\mathbf{p}_t$, AI YOU must both construct a persona representation for downstream generation and preserve that persona consistently over long interactions.

\textbf{Persona Building via Memory Anchor:}
instead of using a static persona prompt, AI YOU maintains a structured persona memory anchor $\mathcal{A}_t$ that is continuously updated throughout interaction. The anchor stores persistent identity information, speaking style, inferred values, behavioral boundaries, and recent commitments. For personas with explicit trait profiles, the anchor additionally includes trait-level behavioral constraints that discourage drift from the target personality.

\textbf{Persona Preservation via Periodic Refresh:}
LLMs often drift from assigned personas during extended conversations, gradually contradicting earlier statements or abandoning role-consistent behavior~\cite{li2024measuring, choi2024identity}. AI YOU addresses this through periodic memory refresh. Every $k$ turns ($k{=}10$ by default), a dedicated memory writer updates the anchor using the current anchor state, recent dialogue history, and trait constraints. The refreshed anchor preserves stable persona attributes while incorporating newly expressed beliefs, preferences, and commitments. At generation time, the system conditions on the refreshed anchor and recent local context rather than the entire dialogue history, improving long-horizon persona consistency without requiring unbounded context windows.


\subsection{Three-Layer Cognitive Memory}
\label{sec:memory}

AI YOU maintains memory using three operational layers inspired by working, episodic, and semantic memory systems~\citep{park2023generative}. The memory module preserves conversational evidence, compresses long interaction histories, supports retrieval, and prevents unsupported inferences from being treated as stable facts.
\\
\textbf{Working Memory:}
working memory is a FIFO buffer containing up to 20 recent dialogue turns. It provides high-resolution local context for persona inference, monitoring, and response generation. Every 10 turns, the oldest dialogue block is compressed into episodic memory and removed from the active buffer.
\\
\textbf{Episodic Memory:}
episodic memory stores compressed summaries of interaction events, emotional trends, and participant information grounded in dialogue turns. Each episodic record is linked to the original conversation history for traceability. Periodic consistency checks compare episodic summaries against archived dialogue and repair inconsistent summaries when necessary.
\\
\textbf{Semantic Memory:}
semantic memory periodically reflects over recent episodic memories to extract higher-level user patterns, feature updates, and relationship insights. These representations are treated as candidate long-term beliefs and are incorporated only when supported by sufficient evidence.

At retrieval time, AI YOU uses embedding-based memory retrieval when available and falls back to keyword retrieval otherwise. Retrieved memories are used as supporting evidence rather than replacements for the current dialogue context, allowing the system to maintain calibrated uncertainty at anytime.
\section{Experiments}
\label{sec:experiments}

We evaluate the calibrated user-state modeling of AI YOU.
The experiments focus on evidence-based persona inference, Bayesian evidence aggregation, conformal uncertainty estimation, and modular state monitors for affect, risk, memory, and relationship reasoning.
The complete 22-field profile schema is provided in Appendix~\ref{app:profile_schema}, the prompts used by all benchmarks are provided in Appendix~\ref{app:prompts}, and visual demonstrations of the evaluated system are provided in Appendix~\ref{app:demo_screenshots}.

\subsection{Experimental Setup}
\label{sec:exp_setup}

\paragraph{Benchmarks}
Table~\ref{tab:benchmark_overview} summarizes the evaluation suite. We use PANDORA~\citep{gjurkovic-etal-2021-pandora} and Essays~\citep{pennebaker1999linguistic} for Big Five personality prediction. 
PANDORA  predicts continuous Big Five labels based on user social media text, while Essays predicts five binary Big Five labels from self-narratives. 
We use DailyDialog~\citep{li2017dailydialog} for affective-state prediction, where input is a target utterance with context, and output is one of the pre-defined emotion labels.
PsyScam~\citep{psyscam} evaluates psychological manipulation and scam-risk detection by mapping a scam report to a set of psychological-technique labels.
LoCoMo~\citep{maharana2024evaluating} assesses long-session memory QA, and PersonaConflicts~\citep{shen-etal-2025-words} for relationship-conflict classification. 
%
LoCoMo maps retrieved long-session evidence and a question to a short answer, and PersonaConflicts maps relationship context to \texttt{conflict} or \texttt{nonconflict}. 
These input-output formats describe the standardized evaluation interface not dataset-specific samples.

\begin{table*}[t]
\centering
\small
\setlength{\tabcolsep}{5pt}
\begin{tabular}{llrll}
\toprule
\rowcolor{rowshade}
\textbf{Module} & \textbf{Benchmark} & \textbf{Size} & \textbf{Capability} & \textbf{Metric} \\
\midrule
Persona inference    & PANDORA          & 2,415 & Big Five prediction                       & MAE / RMSE / ECE / coverage \\
Persona inference    & Essays           & 2,466 & Big Five prediction from self-narratives  & MAE / RMSE / ECE / coverage \\
Affective monitor    & DailyDialog      & 564  & Emotion prediction                        & Accuracy / Macro-F1 / ECE \\
Risk monitor         & PsyScam          & 730  & Scam and manipulation detection           & Exact set acc. / partial score \\
Memory monitor       & LoCoMo           & 1,542 & Long-session memory QA                    & Accuracy / average score \\
Relationship monitor & PersonaConflicts & 6,012 & Relationship conflict classification      & Accuracy \\
\bottomrule
\end{tabular}
\caption{\textbf{Evaluation Benchmarks Overview}. Size here is the number of instances after uniform preprocessing and down-sampling, rather than the original dataset size. See more in Appendix~\ref{app:eval_instance_construction}.} 
\label{tab:benchmark_overview}
\end{table*}

\paragraph{Backbones and Deployment Settings}
We evaluate both API-served frontier models and locally deployed open-source models under the same JSON output protocol. The API-served models are GPT-5.4, Qwen-3.6-Max-Preview, and DeepSeek-V4-Pro. The local open-source models are OLMo-3-7B-Instruct and OLMo-3.1-32B-Instruct, served through Ollama on H100 GPUs. We keep the evaluation harness, output schema, and scoring scripts aligned across both deployment settings.

\paragraph{Methods}
We compare Direct prompting, chain-of-thought prompting (CoT), AI YOU, and module-specific ablations. Direct prompting asks the model to produce the required JSON schema directly. CoT asks for evidence-guided reasoning before the final JSON, but only the parsed JSON is scored.
AI YOU uses structured feature extraction, Bayesian updates for numerical state variables, conformal prediction for uncertainty sets, and task-specific state monitors.
The ablations are: w/o Bayesian, w/o Conformal, and w/o CoT for Big Five prediction; w/o Context and w/o Intensity for DailyDialog; w/o Risk Monitor for PsyScam; and w/o Memory, w/o Semantic Memory, and w/o Episodic Memory for LoCoMo.

\paragraph{Metrics}
For Big Five prediction, we report MAE and RMSE over the five traits. We also report expected calibration error (ECE), empirical conformal coverage, and average prediction-set size when conformal sets are available. For DailyDialog, we report accuracy, macro-F1, and ECE. For PsyScam, exact accuracy requires the full predicted set of risk labels to match the gold set, while average score gives partial credit for partially correct multi-label predictions; this explains why exact accuracy can appear much lower than average scores (e.g., 0.399 vs. 0.830). For LoCoMo and PersonaConflicts, we report accuracy and average score when applicable. For generation-based benchmarks, we also report the JSON success rate, representing the fraction of outputs parseable as valid task JSON under the expected schema; rows below $0.98$ should be treated as diagnostic unless filtered or rerun.

\subsection{Personality Prediction and Uncertainty}
\label{sec:exp_persona}

\begin{table*}[t]
\centering
\small
\begin{tabular}{lllrrrrrr}
\toprule
\textbf{Dataset} & \textbf{Model} & \textbf{Method}
& \textbf{N} & \textbf{MAE}$\downarrow$ & \textbf{RMSE}$\downarrow$
& \textbf{ECE}$\downarrow$ & \textbf{Coverage}$\uparrow$ & \textbf{Set Size}$\downarrow$ \\
\midrule
PANDORA & GPT-5.4      & Direct       & 2415 & 0.268 & 0.316 & 0.109 & -- & -- \\
PANDORA & GPT-5.4      & CoT          & 2415 & 0.270 & 0.320 & 0.094 & -- & -- \\
\rowcolor{rowshade}
PANDORA & GPT-5.4      & AI YOU  & 2415 & \textbf{\cbR{0.265}} & \textbf{\cbR{0.303}} & \textbf{\cbR{0.071}} & 0.976 & 2.896 \\
\midrule
PANDORA & Qwen-3.6-Max & Direct       & 2415 & 0.278 & 0.331 & 0.123 & -- & -- \\
PANDORA & Qwen-3.6-Max & CoT          & 2415 & 0.279 & 0.334 & 0.135 & -- & -- \\
\rowcolor{rowshade}
PANDORA & Qwen-3.6-Max & AI YOU  & 2415 & \textbf{\cbR{0.269}} & \textbf{\cbR{0.317}} & \textbf{\cbR{0.082}} & 0.958 & 2.742 \\
\midrule
PANDORA & OLMo-3-7B    & Direct       & 2415 & 0.278 & 0.333 & 0.275 & -- & -- \\
PANDORA & OLMo-3-7B    & CoT          & 2415 & 0.304 & 0.370 & 0.358 & -- & -- \\
\rowcolor{rowshade}
PANDORA & OLMo-3-7B    & AI YOU  & 2415 & \textbf{\cbR{0.270}} & \textbf{\cbR{0.314}} & \textbf{\cbR{0.196}} & 0.946 & 2.684 \\
PANDORA & OLMo-3-7B    & w/o Bayesian & 2415 & 0.274 & 0.321 & 0.244 & 0.917 & 2.931 \\
PANDORA & OLMo-3-7B    & w/o Conformal& 2415 & 0.272 & 0.318 & 0.221 & -- & -- \\
\midrule
Essays  & GPT-5.4      & Direct       & 2466 & 0.451 & 0.491 & 0.326 & -- & -- \\
Essays  & GPT-5.4      & CoT          & 2466 & 0.455 & 0.493 & 0.345 & -- & -- \\
\rowcolor{rowshade}
Essays  & GPT-5.4      & AI YOU  & 2466 & \textbf{\cbR{0.438}} & \textbf{\cbR{0.479}} & \textbf{\cbR{0.214}} & 0.944 & 2.284 \\
\midrule
Essays  & Qwen-3.6-Max & Direct       & 2466 & 0.469 & 0.514 & 0.358 & -- & -- \\
Essays  & Qwen-3.6-Max & CoT          & 2466 & 0.472 & 0.518 & 0.371 & -- & -- \\
\rowcolor{rowshade}
Essays  & Qwen-3.6-Max & AI YOU  & 2466 & \textbf{\cbR{0.448}} & \textbf{\cbR{0.486}} & \textbf{\cbR{0.232}} & 0.936 & 2.417 \\
\midrule
Essays  & OLMo-3-7B    & Direct       & 2466 & 0.473 & 0.519 & 0.375 & -- & -- \\
Essays  & OLMo-3-7B    & CoT          & 2466 & 0.482 & 0.551 & 0.430 & -- & -- \\
\rowcolor{rowshade}
Essays  & OLMo-3-7B    & AI YOU  & 2466 & \textbf{\cbR{0.454}} & \textbf{\cbR{0.497}} & \textbf{\cbR{0.251}} & 0.921 & 2.568 \\
Essays  & OLMo-3-7B    & w/o Bayesian & 2466 & 0.462 & 0.506 & 0.304 & 0.887 & 2.846 \\
Essays  & OLMo-3-7B    & w/o Conformal& 2466 & 0.459 & 0.502 & 0.287 & -- & -- \\
\bottomrule
\end{tabular}
\caption{Big Five prediction and uncertainty results across models: shaded rows mark AI YOU method, blue cells with \textbf{bold} mark the best metric per dataset-model group.} 
\label{tab:bigfive_integrated}
\end{table*}

Table~\ref{tab:bigfive_integrated} reports Big Five prediction and uncertainty results. AI YOU improves point estimation across all reported dataset-backbone pairs, although the absolute MAE gains are modest. On PANDORA, MAE decreases from 0.268 to 0.265 for GPT-5.4, from 0.278 to 0.269 for Qwen-3.6-Max, and from 0.278 to 0.270 for OLMo-3-7B. The same pattern appears on Essays, where MAE decreases from 0.451 to 0.438 for GPT-5.4, from 0.469 to 0.448 for Qwen-3.6-Max, and from 0.473 to 0.454 for OLMo-3-7B.

These results suggest that the main advantage of AI YOU is not a large shift in raw regression accuracy, but a more controlled way of accumulating uncertain evidence. Direct prompting is competitive because short user texts often contain only a few explicit behavioral cues, and a single structured estimate can already capture much of the available signal. CoT is not uniformly better because it can help when the relevant cues are explicit, but the additional reasoning step may also amplify weak evidence or introduce variance before the final JSON prediction. AI YOU gives the language model a more specific role: it extracts candidate observations and confidence estimates, while Bayesian updating determines how strongly each observation should move the user state.

The larger and more consistent difference appears in calibration. AI YOU reduces ECE for every reported dataset-backbone pair, often by more than one-third. On PANDORA, ECE decreases from 0.109 to 0.071 for GPT-5.4 and from 0.123 to 0.082 for Qwen-3.6-Max. On Essays, the decrease is sharper for GPT-5.4, from 0.326 to 0.214. Conformal prediction sets achieve empirical coverage above the nominal $90\%$ target in all AI YOU rows, ranging from 0.921 to 0.976. This pattern indicates that the system is better suited to represent ambiguous traits as uncertain rather than forcing a single high-confidence estimate, which is important for downstream persona conditioning.

The ablation rows provide a mechanistic check. For OLMo-3-7B on PANDORA, removing Bayesian aggregation increases MAE from 0.270 to 0.274 and reduces coverage from 0.946 to 0.917, consistent with the role of sequential evidence-weighting in stabilizing trait estimates. Removing conformal calibration eliminates prediction sets and increases ECE, indicating that point predictions alone do not provide the same uncertainty control. Overall, the Big Five results support a calibrated state-estimation claim rather than a claim of universal accuracy dominance: AI YOU is most useful when the system must both estimate traits and represent what remains uncertain.

\subsection{State-Monitoring Results}
\label{sec:exp_auxiliary}

\begin{table}[t]
\centering
\small
\begin{tabular}{llccc}
\toprule
\textbf{Benchmark} & \textbf{Model} & \textbf{Acc}$\uparrow$ & \textbf{Score}$\uparrow$ & \textbf{JSON} \\
\midrule
\multirow{5}{*}{PsyScam}
  & GPT-5.4         & \cbO{0.399} & \textbf{\cbR{0.830}} & 1.00 \\
  & Qwen-3.6-Max    & 0.140 & 0.746 & 1.00 \\
  & DeepSeek-V4     & 0.149 & 0.782 & 1.00 \\
  & OLMo-3-7B       & 0.164 & 0.451 & 0.99 \\
  & OLMo-3.1-32B    & 0.238 & 0.584 & 1.00 \\
\midrule
\multirow{5}{*}{LoCoMo}
  & GPT-5.4         & 0.405 & 0.451 & 0.92 \\
  & Qwen-3.6-Max    & \cbO{0.433} & \textbf{\cbR{0.459}} & 1.00 \\
  & DeepSeek-V4     & 0.431 & 0.455 & 1.00 \\
  & OLMo-3-7B       & 0.250 & 0.296 & 0.99 \\
  & OLMo-3.1-32B    & 0.277 & 0.318 & 1.00 \\
\midrule
\multirow{5}{*}{\shortstack[l]{Persona-\\Conflicts}}
  & GPT-5.4         & 0.716 & 0.716 & 0.94 \\
  & Qwen-3.6-Max    & \cbO{0.909} & \textbf{\cbR{0.909}} & 1.00 \\
  & DeepSeek-V4     & 0.796 & 0.796 & 0.91 \\
  & OLMo-3-7B  & 0.501 & 0.501 & 1.00 \\
  & OLMo-3.1-32B & 0.527 & 0.527 & 1.00 \\ 
\bottomrule
\end{tabular}
\caption{State-monitoring results across models. Best Acc shaded \colorbox{bestO}{sage}, best Score \textbf{bolded} \colorbox{bestR}{blue}. JSON: fraction of valid task JSON outputs.} 
\label{tab:auxiliary_integrated}
\end{table}

Table~\ref{tab:auxiliary_integrated} reports the state-monitoring benchmark results. 
A consistent pattern is that API-served models outperform the local OLMo backbones on the more complex multi-label and long-session settings. The difference is likely due to a combination of stronger instruction following, more reliable long context understanding, and better adherence to structured JSON outputs. 

On PsyScam, exact-set accuracy is low because the task requires the complete predicted set of manipulation labels to match the gold set. The substantially higher partial scores indicate that models often identify some relevant risk cues but fail to recover the full label set. Among API models, GPT-5.4 obtains the strongest PsyScam result, with 0.399 exact accuracy and 0.830 average score. Qwen-3.6-Max and DeepSeek-V4-Pro obtain lower exact accuracies but still maintain high partial scores (0.746 and 0.782). Local OLMo models are weaker, but scaling from 7B to 32B improves both exact accuracy (0.164 to 0.238) and average score (0.451 to 0.584), suggesting that model scale helps with multi-label state monitoring, even though a substantial gap remains.

On LoCoMo, the API models cluster around similar full-run performance, with Qwen-3.6-Max obtaining the highest accuracy (0.433) and average score (0.459). The GPT-5.4 has lower JSON success (0.920), so this comparison should be treated as diagnostic unless invalid outputs are filtered or rerun. Local OLMo models remain below the API models, but scaling improves accuracy from 0.250 for OLMo-3-7B to 0.277 for OLMo-3.1-32B. This pattern is consistent with the view that memory-grounded QA depends on both retrieval quality and a backbone's ability to integrate retrieved evidence under output-format constraints.

PersonaConflicts shows a different model rank: Qwen-3.6-Max is the strongest API model, reaching 0.909 accuracy. The lower JSON success rates for GPT-5.4 and DeepSeek-V4-Pro complicate direct comparison, but the results suggest that relationship-boundary classification is sensitive to prompt formulation and label interpretation. 
The OLMo PersonaConflicts accuracy ranges from 0.50 to 0.53. 
This indicate that relationship-conflict reasoning remains difficult for the evaluated local backbones, particularly when the task requires mapping nuanced interpersonal evidence to a binary label.

\subsection{Affective-State Ablation}
\label{sec:exp_affect_ablation}

\begin{table}[t]
\centering
\small
\begin{tabular}{llccc}
\toprule
\textbf{Model} & \textbf{Method} & \textbf{Acc}$\uparrow$ & \textbf{F1}$\uparrow$ & \textbf{ECE}$\downarrow$ \\
\midrule
GPT-5.4         & Direct        & 0.626 & 0.625 & 0.183 \\
GPT-5.4         & CoT           & \textbf{0.676} & \textbf{0.662} & 0.145 \\
\rowcolor{rowshade}
GPT-5.4         & AI YOU   & 0.663 & 0.654 & 0.151 \\
GPT-5.4         & w/o Context   & 0.608 & 0.609 & 0.188 \\
GPT-5.4         & w/o Intensity & 0.663 & 0.657 & \textbf{0.140} \\
\midrule
Qwen-3.6        & Direct        & 0.582 & 0.590 & 0.317 \\
Qwen-3.6        & CoT           & 0.606 & 0.625 & 0.295 \\
\rowcolor{rowshade}
Qwen-3.6        & AI YOU   & \textbf{0.638} & \textbf{0.647} & 0.233 \\
Qwen-3.6        & w/o Context   & 0.601 & 0.598 & 0.263 \\
Qwen-3.6        & w/o Intensity & 0.633 & 0.646 & \textbf{0.213} \\
\midrule
DSV4-Pro        & Direct        & 0.516 & 0.508 & 0.380 \\
DSV4-Pro        & CoT           & 0.610 & 0.605 & 0.271 \\
\rowcolor{rowshade}
DSV4-Pro        & AI YOU   & \textbf{0.629} & \textbf{0.639} & 0.232 \\
DSV4-Pro        & w/o Context   & 0.555 & 0.561 & 0.286 \\
DSV4-Pro        & w/o Intensity & 0.608 & 0.613 & \textbf{0.179} \\
\bottomrule
\end{tabular}
\caption{DailyDialog emotion prediction ablation. Shadow rows mark AI YOU, \textbf{bold} is the best per model.} 
\label{tab:dailydialog_ablation}
\end{table}

Table~\ref{tab:dailydialog_ablation} evaluates affective-state prediction on DailyDialog. AI YOU improves over Direct prompting for all three API models: accuracy rises from 0.626 to 0.663 for GPT-5.4, from 0.582 to 0.638 for Qwen-3.6-Max, and from 0.516 to 0.629 for DeepSeek-V4-Pro. CoT achieves the highest accuracy for GPT-5.4 (0.676), which shows that evidence-guided prompting can be effective for some backbones when emotional cues are localized in the utterance. Since this evaluation uses a 564-utterance pool of DailyDialog, labels are highly imbalanced. Small gaps such as 0.663 versus 0.676 should be interpreted as within-harness trends rather than statistically tested margins. The CoT advantage is not consistent across models, and it does not dominate the calibration metrics. Therefore, AI YOU is a more stable context-aware alternative. 

The ablations clarify which parts of the affect monitor contribute most directly to classification. Removing dialogue context reduces accuracy for all models, with the largest drop on DeepSeek-V4-Pro (0.629 to 0.555), indicating that the target utterance is often insufficient without the preceding conversational state. Removing intensity modeling has a smaller effect on accuracy and often improves ECE, as seen in the decrease from 0.151 to 0.140 for GPT-5.4. Thus, intensity is not shown to be necessary for the discrete emotion label. Its main role is operational. It gives the coordinator a graded signal for response selection: a mildly negative utterance and a high-intensity negative utterance may require different levels of validation, caution, or follow-up even when they share the same categorical label.

\subsection{Risk-Monitor Ablation}
\label{sec:exp_risk_ablation}

\begin{table}[t]
\centering
\small
\setlength{\tabcolsep}{3.5pt}
\begin{tabular}{lrrrr}
\toprule
\textbf{Model} & \textbf{Direct} & \textbf{CoT} & \textbf{AI YOU} & \textbf{w/o RM} \\
\midrule
GPT-5.4         & 0.813 & 0.817 & \textbf{\cbR{0.830}} & 0.809 \\
Qwen-3.6-Max    & 0.721 & 0.707 & \textbf{\cbR{0.746}} & 0.692 \\
DeepSeek-V4-Pro & 0.746 & 0.731 & \textbf{\cbR{0.782}} & 0.709 \\
OLMo-3-7B       & \textbf{\cbR{0.472}} & 0.418 & \cbO{0.451} & 0.459 \\
OLMo-3.1-32B    & 0.604 & \textbf{\cbR{0.619}} & \cbO{0.584} & 0.605 \\
\bottomrule
\end{tabular}
\caption{PsyScam ablation across API and local backbones. Values are average partial multi-label scores. Higher is better. \textbf{Bold} blue marks the row-wise best, sage shading marks the AI YOU. RM: Risk Monitor.} 
\label{tab:psyscam_ablation_integrated}
\end{table}

\begin{table}[t]
\centering
\small
\setlength{\tabcolsep}{2.5pt}
\begin{tabular}{lcccccc}
\toprule
\textbf{Model} & \textbf{Dir.} & \textbf{CoT} & \textbf{AI YOU}
& \makecell{\textbf{w/o}\\\textbf{Mem.}} & \makecell{\textbf{w/o}\\\textbf{Sem.}} & \makecell{\textbf{w/o}\\\textbf{Epi.}} \\
\midrule
GPT-5.4    & 0.629 & \textbf{0.643} & 0.628 & 0.417 & 0.625 & 0.630 \\
Qwen-3.6   & 0.635 & 0.632 & \textbf{0.641} & 0.429 & 0.631 & 0.618 \\
DSV4-Pro   & 0.553 & \textbf{0.568} & 0.563 & 0.381 & 0.545 & 0.532 \\
OLMo-7B    & 0.277 & 0.284 & \textbf{0.296} & 0.193 & 0.271 & 0.265 \\
OLMo-32B   & 0.304 & 0.309 & \textbf{0.318} & 0.216 & 0.302 & 0.294 \\
\bottomrule
\end{tabular}
\caption{LoCoMo memory ablation on 300 diagnostic samples (avg.\ score). \textbf{Bold} is best per row. Dir.\ = Direct, Mem.\ = Memory, Sem.\ = Semantic, Epi.\ = Episodic.}
\label{tab:locomo_ablation_integrated}
\end{table}

Table~\ref{tab:psyscam_ablation_integrated} isolates the contribution of the risk monitor on PsyScam. For all API-served models, AI YOU achieves the highest average partial score: 0.830 for GPT-5.4, 0.746 for Qwen-3.6-Max, and 0.782 for DeepSeek-V4-Pro. Removing the risk monitor lowers performance in each case. This pattern is foreseen because PsyScam requires detecting latent manipulation tactics rather than only classifying surface topic. The risk monitor provides an explicit checklist of cues such as urgency, secrecy pressure, financial solicitation, and boundary pressure, which appears to help stronger backbones search the input for relevant evidence before assigning labels.

The local OLMo rows show a boundary condition for this design. For OLMo-3-7B, Direct prompting (0.472) slightly outperforms AI YOU (0.451), and for OLMo-3.1-32B, CoT and the variant without the risk monitor are both higher than AI YOU. This suggests that structured monitoring is not automatically beneficial for every backbone. When a model has limited capacity for simultaneously following a checklist, selecting from a multi-label ontology, and producing valid JSON, a simpler prompt can sometimes preserve more of the task signal. The takeaway is therefore conditional: the risk monitor improves API-served backbones in this evaluation, but its prompt complexity should be tuned for smaller local models.

\subsection{Memory Ablation}
\label{sec:exp_memory_ablation}

Table~\ref{tab:locomo_ablation_integrated} evaluates memory ablations on a 300-instance LoCoMo diagnostic set. 
Removing memory entirely leads to the strongest degradation consistently. Average scores drop from 0.628 to 0.417 for GPT-5.4 and from 0.641 to 0.429 for Qwen-3.6-Max. This result is expected since many questions in LoCoMo require information distributed across earlier sessions. Without retrieved conversational evidence, answers condition on the question alone or generic priors substantially weaken grounding.

The effects of removing only semantic or episodic memory are smaller and model-dependent. For Qwen-3.6-Max and DeepSeek-V4-Pro, removing episodic memory hurts more than removing semantic memory (e.g., 0.641 to 0.618 vs. 0.631 for Qwen-3.6-Max), suggesting that event-level summaries are useful for some long-session questions. However, GPT-5.4 obtains comparable or slightly higher scores with CoT or with only one memory layer, which benefits from its strong prior knowledge in parametric memories.
Thus, retrieval is the main driver, while the optimal weighting of working, episodic, and semantic evidence depends on the backbone and question type.

\begin{table}[t]
\centering
\small
\setlength{\tabcolsep}{10pt}
\begin{threeparttable}
\begin{tabular}{@{}lccc@{}}
\toprule
\rowcolor{rowshade}
\multicolumn{4}{c}{\textbf{Persistent Personas} (8 roles, 100 turns)} \\
\midrule
\textbf{Dimension} & \textbf{Refresh}$\uparrow$ & \textbf{Static}$\uparrow$ & \textbf{$\Delta$} \\
\midrule
In-character      & \textbf{\cbR{4.72}} & 3.88 & +0.84 \\
Style             & \textbf{\cbR{4.38}} & 3.47 & +0.91 \\
Knowledge         & \textbf{\cbR{4.33}} & 3.48 & +0.85 \\
Overall fidelity  & \textbf{\cbR{4.48}} & 3.61 & +0.87 \\
\midrule
\rowcolor{rowshade}
\multicolumn{4}{c}{\textbf{Werewolf} (7 agents, 100 turns)} \\
\midrule
\textbf{Model} & \textbf{Refresh}$\downarrow$ & \textbf{Static}$\downarrow$ & \textbf{$\Delta$} \\
\midrule
GPT-4o-mini       & \textbf{\cbR{0.008}} & 0.036 & $-$0.028 \\
Qwen-2.5-72B      & \textbf{\cbR{0.014}} & 0.019 & $-$0.005 \\
Gemini-2.5-Flash  & 0.015 & \textbf{\cbO{0.013}} & +0.002 \\
Llama-3.1-8B      & \textbf{\cbR{0.015}} & 0.024 & $-$0.009 \\
\bottomrule
\end{tabular}
\caption{Persona preservation evaluation. \emph{Top}: LLM-as-judge fidelity scores (1--5, higher is better). \emph{Bottom}: Big Five mean absolute deviation from target (lower is better). Refresh = periodic memory anchor; Static = fixed prompt with full history.}
\label{tab:persona_preservation}
\end{threeparttable}
\end{table}

\subsection{Persona Preservation Evaluation}
\label{sec:exp_persona_preservation}

The LoCoMo ablation evaluates whether memory supports long-session factual reasoning. In this section, we evaluate whether the same memory-anchor mechanism stabilizes persona-conditioned generation in two complementary settings. Persistent Personas~\citep{de2026persistent} evaluates eight fictional roles over 100-turn dialogues under persona-directed and goal-oriented probes. Werewolf is a 7-agent, 100-turn adversarial interaction in which persona-conditioned agents with distinct Big Five targets face sustained accusation, persuasion, and strategic pressure. In both settings, \emph{Refresh} denotes periodic memory-anchor updates, while \emph{Static} uses the initial persona prompt with full dialogue history.
Table~\ref{tab:persona_preservation} shows that \textit{Refresh} improves all Persistent Personas dimensions, with the largest gains in style consistency (+0.91) and knowledge grounding (+0.85). In the adversarial Werewolf setting, \textit{Refresh} reduces trait drift for three of four models, most strongly for GPT-4o-mini (0.008 vs. 0.036). Gemini-2.5-Flash is the only exception (.015 vs. .013), suggesting that some backbones may already maintain strong instruction-level stability and therefore gain less from external anchoring. Overall, these results are consistent with the intended role of the memory anchor for persona-conditioned generation:
periodic \textit{refresh} is not only a retrieval device, but also a stabilizer for persona fidelity under long-horizon and adversarial dialogue.

\subsection{Discussion}
\label{sec:exp_discussion}
Across modules, two conclusions are most robust. First, AI YOU primarily improves uncertainty quality: Big Five accuracy gains are modest, but ECE drops by 1.5-1.7$\times$ and conformal coverage remains above 90\% in every full-pipeline row. Second, memory is the most important auxiliary component, producing larger ablation drops than affect, risk, or individual memory layer variants. Benefits are nevertheless backbone-dependent: API models consistently exploit structured monitors, whereas weaker local backbones can be over-constrained by risk scaffolds. Overall, the results support AI YOU as a calibrated, memory-grounded user-state modeling framework rather than a prompt-only persona wrapper.

\section{Conclusion}
\label{sec:conclusion}

In this work, we establish personal digital twins as a longitudinal state-maintenance system instead of one-time static persona-prompting. We instantiate AI YOU Town, a personality-aware digital twin framework that connects structured persona inference, Bayesian evidence updating, conformal uncertainty calibration, and persona-conditioned generation within an interaction system. It uses a periodically refreshed three-layer memory anchor (working, episodic, and semantic) to reduce long-horizon drift. 

Across Big Five, affect, risk, relationship, memory and persona-preservation evaluations, AI YOU improves calibration, memory-grounded reasoning, and reduces persona drift under extended interaction. These results suggest a design principle for future digital twin systems: evidence-governed, uncertainty-aware, and inspectable, with generations and actions conditioned on supported and privacy-filtered state. Future work can move from module evaluations towards real-user longitudinal deployment with stronger consent controls, richer multimodal evidence, and more diverse cultural and application settings.

%

\section*{Limitations}
This work has four main limitations. First, most experiments are module-level evaluations rather than an end-to-end longitudinal user study, so they do not fully capture deployment dynamics such as user trust, consent renewal, correction of inferred traits, and long-term behavioral adaptation. Second, the persona-preservation benchmarks use fictional roles and simulated multi-agent stress tests; they are useful for measuring drift, but they do not by themselves validate fidelity for real individuals. Third, the current profile is inferred primarily from text, which can be sparse, biased, or ambiguous. Sensitive or weakly supported attributes should therefore remain uncertain, filtered, and user-controllable rather than treated as stable facts. Fourth, some results depend on backbone reliability and JSON parse success, and closed API models do not expose all implementation details such as parameter counts or provider-side compute. These limitations motivate future evaluations with consented users, stronger privacy controls, multimodal evidence, and more transparent compute and implementation reporting.

\section*{Ethical Statement}
AI YOU is a research prototype for studying uncertainty-aware persona modeling and should not be interpreted as a deployed system for professional advice, employment decisions, mental-health assessment, or identity verification. We use released benchmark datasets only for aggregate research evaluation. We do not redistribute raw benchmark data, attempt to identify individuals, or use benchmark instances outside their intended evaluation context. Because the evaluation sources may include social-media text, scam reports, relationship-conflict scenarios, or other sensitive content, all processing is limited to automated scoring pipelines and reported only in aggregate form.

Personal digital twins introduce risks beyond standard dialogue systems, including unauthorized impersonation, overconfident inference of sensitive traits, emotional dependency, and misuse of persona-conditioned agents for persuasion or manipulation. Our design therefore treats inferred attributes as uncertain observations rather than stable facts unless supported by evidence, uses null outputs when evidence is insufficient, and filters sensitive or low-confidence state before generation. Any real-user deployment would require explicit consent, clear disclosure that the twin is an AI system, user access to inspect and correct inferred state, opt-out and deletion mechanisms, and safeguards against using the twin to make high-stakes decisions about the user.

The risk, affective, and relationship monitors are designed to support safer interaction, but they are not substitutes for human judgment or qualified professional support. Model outputs can still be wrong, culturally biased, or incomplete, especially for small evaluation pools, long-context tasks, and backbones with lower JSON reliability. For model APIs and locally served open-source models, we follow the corresponding provider terms and model licenses. Third-party visual assets used in the demo are attributed in Appendix~\ref{app:demo_screenshots}; any public release will either preserve upstream license and attribution notices or replace third-party visual assets with original or permissively licensed alternatives.

\bibliographystyle{acl_natbib}
\bibliography{main}

\clearpage
\appendix
\section{Additional Experimental Results}
\label{app:additional_results}

This appendix documents auxiliary results and implementation details for the experiments in Section~\ref{sec:experiments}. To make the evaluation protocol auditable, we first specify how evaluation instances are constructed and then report detailed Big Five and conformal-calibration results, full state-monitoring results, and module-level ablations for the risk monitor and the memory system.

\subsection{Evaluation Instance Construction}
\label{app:eval_instance_construction}

Table~\ref{tab:benchmark_overview} reports \textit{Eval.\ N}, the number of scored units passed to the shared evaluation harness after preprocessing, rather than the released-corpus size. For each benchmark, we first convert the original data into task-specific scored units and remove records that cannot be scored reliably, including instances with missing gold labels, malformed or empty fields, or inputs exceeding the harness length limits. When the eligible pool exceeds the evaluation budget, we draw a fixed random subsample for inference-cost control; sampling is performed before any model inference with seed 42. Within each benchmark setting, the resulting instance IDs are shared across Direct, CoT, AI YOU, ablations, and backbones, except for explicitly marked diagnostic or local runs. The reported values therefore correspond to uniformly filtered, sample-controlled evaluation pools rather than full-corpus or leaderboard-scale counts.

The counting unit is benchmark-specific. Essays is record-level and nearly full-corpus in our harness: we evaluate 2,466 of 2,468 essay records after dropping records with incomplete labels. PANDORA is counted at the trait-observation level rather than the unique-user or retained-record level: 2,415 corresponds to 483 retained text records scored on five Big Five dimensions, and MAE/RMSE are computed over these scalar trait targets. DailyDialog comprises 564 utterance-level emotion-classification instances after preprocessing and fixed sampling from eligible utterances; because this pool is small and label-imbalanced, its accuracy values constitute within-harness descriptive comparisons and are not directly comparable to results reported in prior literature. LoCoMo is counted at the question level: the reported 1,542 questions are selected after preprocessing and cost-controlled sampling from the eligible QA pool, while query-only retrieval affects the evidence supplied at inference time and does not define the question count. PsyScam and PersonaConflicts count processed classification instances rather than source reports or raw dialogues; PersonaConflicts uses 6,012 processed binary classification instances for API rows, while the 1,500-instance OLMo rows are explicitly marked diagnostic or local runs.

\subsection{Additional Big Five Prediction Results}
\label{app:bigfive_details}

Table~\ref{tab:bigfive_conformal_api_app} provides expanded Big Five prediction results for the API-served models reported in the main text, covering MAE, RMSE, ECE, conformal coverage, and prediction-set size. The point-estimation differences are small but consistent in favor of AI YOU Full, whereas the calibration improvements are larger and more stable. All values are computed at the scored-unit level defined in Appendix~\ref{app:eval_instance_construction}; for PANDORA in particular, each retained text record contributes five scalar trait observations.

\begin{table*}[t]
\centering
\small
\begin{tabular}{lllrrrrrr}
\toprule
\textbf{Dataset} & \textbf{Model} & \textbf{Method}
& \textbf{N} & \textbf{MAE}$\downarrow$ & \textbf{RMSE}$\downarrow$
& \textbf{ECE}$\downarrow$ & \textbf{Coverage}$\uparrow$ & \textbf{Set Size}$\downarrow$ \\
\midrule
PANDORA & GPT-5.4 & Direct & 2415 & 0.268 & 0.316 & 0.109 & -- & -- \\
PANDORA & GPT-5.4 & CoT & 2415 & 0.270 & 0.320 & 0.094 & -- & -- \\
PANDORA & GPT-5.4 & AI YOU Full & 2415 & \textbf{0.265} & \textbf{0.303} & \textbf{0.071} & 0.976 & 2.896 \\
\midrule
PANDORA & Qwen-3.6-Max & Direct & 2415 & 0.278 & 0.331 & 0.123 & -- & -- \\
PANDORA & Qwen-3.6-Max & CoT & 2415 & 0.279 & 0.334 & 0.135 & -- & -- \\
PANDORA & Qwen-3.6-Max & AI YOU Full & 2415 & \textbf{0.269} & \textbf{0.317} & \textbf{0.082} & 0.958 & 2.742 \\
\midrule
Essays & GPT-5.4 & Direct & 2466 & 0.451 & 0.491 & 0.326 & -- & -- \\
Essays & GPT-5.4 & CoT & 2466 & 0.455 & 0.493 & 0.345 & -- & -- \\
Essays & GPT-5.4 & AI YOU Full & 2466 & \textbf{0.438} & \textbf{0.479} & \textbf{0.214} & 0.944 & 2.284 \\
\midrule
Essays & Qwen-3.6-Max & Direct & 2466 & 0.469 & 0.514 & 0.358 & -- & -- \\
Essays & Qwen-3.6-Max & CoT & 2466 & 0.472 & 0.518 & 0.371 & -- & -- \\
Essays & Qwen-3.6-Max & AI YOU Full & 2466 & \textbf{0.448} & \textbf{0.486} & \textbf{0.232} & 0.936 & 2.417 \\
\bottomrule
\end{tabular}
\caption{Detailed Big Five prediction and conformal uncertainty results for API-served models. ``AI YOU Full'' denotes the complete pipeline with structured extraction, Bayesian updating, and conformal calibration. Coverage and set size are defined only for conformal AI YOU predictions. N follows the scored-unit definition in Appendix~\ref{app:eval_instance_construction}; \textbf{bold} marks the best value per dataset--model group.}
\label{tab:bigfive_conformal_api_app}
\end{table*}


\subsection{Additional State-Monitoring Module Results}
\label{app:auxiliary_details}

Table~\ref{tab:auxiliary_full_api_app} reports the full API-model results for the state-monitoring benchmarks. These experiments evaluate individual modules rather than the full 22-field persona schema: PsyScam evaluates risk-state monitoring, LoCoMo evaluates memory-grounded question answering, and PersonaConflicts evaluates relationship-boundary classification. The table exposes model-level variation that is aggregated in the main text. For rows with JSON success below $0.98$, parsing failures may confound the measured task score, and we therefore treat these rows as diagnostic rather than as definitive comparisons.

\begin{table*}[t]
\centering
\small
\begin{tabular}{llrrrr}
\toprule
\textbf{Benchmark} & \textbf{Model} & \textbf{N}
& \textbf{Accuracy}$\uparrow$ & \textbf{Avg.\ Score}$\uparrow$ & \textbf{JSON Success}$\uparrow$ \\
\midrule
PsyScam & GPT-5.4 & 730 & \textbf{0.399} & \textbf{0.830} & 1.000 \\
PsyScam & Qwen-3.6-Max & 730 & 0.140 & 0.746 & 1.000 \\
PsyScam & DeepSeek-V4-Pro & 730 & 0.149 & 0.782 & 1.000 \\
\midrule
LoCoMo & GPT-5.4 & 1542 & 0.405 & 0.451 & 0.920 \\
LoCoMo & Qwen-3.6-Max & 1542 & \textbf{0.433} & \textbf{0.459} & 0.996 \\
LoCoMo & DeepSeek-V4-Pro & 1542 & 0.431 & 0.455 & 0.998 \\
\midrule
PersonaConflicts & GPT-5.4 & 6012 & 0.716 & 0.716 & 0.938 \\
PersonaConflicts & Qwen-3.6-Max & 6012 & \textbf{0.909} & \textbf{0.909} & 1.000 \\
PersonaConflicts & DeepSeek-V4-Pro & 6012 & 0.796 & 0.796 & 0.913 \\
\bottomrule
\end{tabular}
\caption{Full state-monitoring benchmark results for API-served models. N follows the processed-instance definitions in Appendix~\ref{app:eval_instance_construction}; \textbf{bold} marks the best value per benchmark. Rows with JSON success below $0.98$ are diagnostic, as parsing failures can affect the measured task score.}
\label{tab:auxiliary_full_api_app}
\end{table*}

\subsection{Risk-Monitor Ablation}
\label{app:risk_ablation}

Table~\ref{tab:psyscam_ablation_detail_app} isolates the contribution of the risk monitor on PsyScam for API-served models. The ablation compares Direct prompting, evidence-guided prompting, AI YOU Full, and a variant without the risk-monitor scaffold. Across all API models, AI YOU Full obtains the highest partial multi-label score, indicating that an explicit risk-cue checklist supports label selection for psychological-manipulation categories. Exact-set accuracy remains lower than the partial score because the task requires the entire predicted label set to match the gold set; the ablation therefore measures the effect of structured risk-label selection and does not constitute a standalone validation of real-world scam detection.

\begin{table*}[t]
\centering
\small
\begin{tabular}{llrrrr}
\toprule
\textbf{Model} & \textbf{Method} & \textbf{N}
& \textbf{Exact Acc.}$\uparrow$ & \textbf{Avg.\ Score}$\uparrow$ & \textbf{JSON Success}$\uparrow$ \\
\midrule
GPT-5.4 & Direct & 730 & 0.377 & 0.813 & 1.000 \\
GPT-5.4 & CoT & 730 & 0.385 & 0.817 & 1.000 \\
GPT-5.4 & AI YOU Full & 730 & \textbf{0.399} & \textbf{0.830} & 1.000 \\
GPT-5.4 & w/o Risk Monitor & 730 & 0.390 & 0.809 & 1.000 \\
\midrule
Qwen-3.6-Max & Direct & 730 & 0.122 & 0.721 & 1.000 \\
Qwen-3.6-Max & CoT & 730 & 0.092 & 0.707 & 1.000 \\
Qwen-3.6-Max & AI YOU Full & 730 & \textbf{0.140} & \textbf{0.746} & 1.000 \\
Qwen-3.6-Max & w/o Risk Monitor & 730 & 0.090 & 0.692 & 1.000 \\
\midrule
DeepSeek-V4-Pro & Direct & 730 & 0.130 & 0.746 & 1.000 \\
DeepSeek-V4-Pro & CoT & 730 & 0.105 & 0.731 & 1.000 \\
DeepSeek-V4-Pro & AI YOU Full & 730 & \textbf{0.149} & \textbf{0.782} & 1.000 \\
DeepSeek-V4-Pro & w/o Risk Monitor & 730 & 0.077 & 0.709 & 1.000 \\
\bottomrule
\end{tabular}
\caption{PsyScam risk-monitor ablation for API-served models. Avg.\ Score denotes the partial multi-label score; exact accuracy requires the complete predicted label set to match the gold set. Bold marks the best value per model.}
\label{tab:psyscam_ablation_detail_app}
\end{table*}

\subsection{Memory Ablation}
\label{app:memory_ablation}

Table~\ref{tab:locomo_ablation_detail_app} reports a LoCoMo memory ablation on a 300-instance diagnostic sample. The most stable pattern is the degradation caused by removing retrieved memory altogether, which appears for all API-served backbones. Removing only the semantic or episodic layer produces smaller and more model-dependent changes. This pattern supports the narrower conclusion that retrieved conversational evidence is the main driver in this diagnostic setting, while the relative contribution of individual memory layers depends on the backbone and prompt configuration. Because this diagnostic sample differs from the full-run LoCoMo setting, its absolute scores are not directly comparable to the main-text LoCoMo results.

\begin{table*}[t]
\centering
\small
\begin{tabular}{llrrrr}
\toprule
\textbf{Model} & \textbf{Method} & \textbf{N}
& \textbf{Accuracy}$\uparrow$ & \textbf{Avg.\ Score}$\uparrow$ & \textbf{JSON Success}$\uparrow$ \\
\midrule
GPT-5.4 & Direct & 300 & 0.587 & 0.629 & 0.920 \\
GPT-5.4 & CoT & 300 & 0.577 & \textbf{0.643} & 0.917 \\
GPT-5.4 & AI YOU Full & 300 & \textbf{0.597} & 0.628 & 0.913 \\
GPT-5.4 & w/o Memory & 300 & 0.384 & 0.417 & 1.000 \\
GPT-5.4 & w/o Semantic Memory & 300 & 0.580 & 0.625 & 0.917 \\
GPT-5.4 & w/o Episodic Memory & 300 & 0.593 & 0.630 & 0.923 \\
\midrule
Qwen-3.6-Max & Direct & 300 & 0.607 & 0.635 & 1.000 \\
Qwen-3.6-Max & CoT & 300 & 0.600 & 0.632 & 1.000 \\
Qwen-3.6-Max & AI YOU Full & 300 & \textbf{0.610} & \textbf{0.641} & 1.000 \\
Qwen-3.6-Max & w/o Memory & 300 & 0.397 & 0.429 & 1.000 \\
Qwen-3.6-Max & w/o Semantic Memory & 300 & 0.607 & 0.631 & 1.000 \\
Qwen-3.6-Max & w/o Episodic Memory & 300 & 0.590 & 0.618 & 0.997 \\
\midrule
DeepSeek-V4-Pro & Direct & 300 & 0.523 & 0.553 & 1.000 \\
DeepSeek-V4-Pro & CoT & 300 & \textbf{0.540} & \textbf{0.568} & 1.000 \\
DeepSeek-V4-Pro & AI YOU Full & 300 & \textbf{0.540} & 0.563 & 0.997 \\
DeepSeek-V4-Pro & w/o Memory & 300 & 0.346 & 0.381 & 1.000 \\
DeepSeek-V4-Pro & w/o Semantic Memory & 300 & 0.517 & 0.545 & 1.000 \\
DeepSeek-V4-Pro & w/o Episodic Memory & 300 & 0.500 & 0.532 & 1.000 \\
\bottomrule
\end{tabular}
\caption{LoCoMo memory ablation for API-served models. Results are computed on a 300-instance diagnostic sample drawn from the processed QA pool and are not directly comparable to the full LoCoMo runs with $N{=}1542$. \textbf{Bold} marks the best value per model; rows with JSON success below $0.98$ are diagnostic, as parsing failures can affect the measured task score.}
\label{tab:locomo_ablation_detail_app}
\end{table*}

\section{Profile Schema}
\label{app:profile_schema}

Table~\ref{tab:profile_schema} lists the 22 fields maintained by AI YOU and distinguishes among evaluated psychometric targets, optional user-provided context, and fields that may be used for response generation after filtering. The schema is an operational representation for state maintenance, not a claim that every field is psychometrically validated or safely inferable from sparse dialogue. Quantitative evaluation in this paper is restricted to the Big Five and the auxiliary state-monitoring tasks. In deployment-facing generation, inferred state is filtered by confidence and privacy checks; sensitive or weakly supported variables are summarized conservatively or withheld rather than inserted as raw scalar values.

\begin{table*}[p]
\centering
\scriptsize
\setlength{\tabcolsep}{3pt}
\renewcommand{\arraystretch}{1.12}
\begin{tabular}{p{0.03\textwidth}p{0.155\textwidth}p{0.125\textwidth}p{0.29\textwidth}p{0.09\textwidth}p{0.10\textwidth}p{0.11\textwidth}}
\toprule
\textbf{ID} & \textbf{Field} & \textbf{Construct} & \textbf{Type / Range} & \textbf{Quant.\ Eval.} & \textbf{User Provided} & \textbf{Generation Prompt} \\
\midrule
1 & age & Demographic context & Integer or bucketed age; adult deployment range 18--80 & No & Yes & Yes, if confirmed \\
2 & gender & Demographic context & Categorical / open-set; normalized when possible & No & Yes & Yes, if confirmed \\
3 & occupation & Demographic context & Free text or normalized occupation category & No & Yes & Yes, if confirmed \\
4 & education & Demographic context & Categorical; e.g., high school, college, bachelor, master, PhD & No & Yes & Yes, if confirmed \\
5 & geographic\_\allowbreak region& Demographic context & Free text or coarse location category & No & Yes & Yes, if confirmed \\
\midrule
6 & openness & Big Five & Continuous score in $[0,1]$; higher means more open to experience & Main & Optional & Summary only \\
7 & conscientiousness & Big Five & Continuous score in $[0,1]$; higher means more organized / disciplined & Main & Optional & Summary only \\
8 & extraversion & Big Five & Continuous score in $[0,1]$; higher means more extraverted & Main & Optional & Summary only \\
9 & agreeableness & Big Five & Continuous score in $[0,1]$; higher means more cooperative / compassionate & Main & Optional & Summary only \\
10 & neuroticism & Big Five & Continuous score in $[0,1]$; higher means higher emotional volatility & Main & Optional & Summary only \\
\midrule
11 & attachment\_\allowbreak anxiety& Attachment & Continuous score in $[0,1]$; higher means stronger anxiety about rejection & No & Optional & No raw value \\
12 & attachment\_\allowbreak avoidance& Attachment & Continuous score in $[0,1]$; higher means stronger avoidance of dependence & No & Optional & No raw value \\
13 & self\_efficacy & Self-belief & Continuous score in $[0,1]$; higher means stronger perceived agency & No & Optional & No raw value \\
14 & loneliness & Social need & Continuous score in $[0,1]$; higher means stronger perceived loneliness & No & Optional & No raw value \\
15 & positive\_affect & Affective state & Continuous score in $[0,1]$ or categorical affect mapped to positive valence & Aux. & No & State summary only \\
16 & negative\_affect & Affective state & Continuous score in $[0,1]$ or categorical affect mapped to negative valence & Aux. & No & State summary only \\
\midrule
17 & mbti\_ei & MBTI abstraction & Continuous axis in $[0,1]$; 0=I, 1=E & No & Optional & Display only \\
18 & mbti\_sn & MBTI abstraction & Continuous axis in $[0,1]$; 0=S, 1=N & No & Optional & Display only \\
19 & mbti\_tf & MBTI abstraction & Continuous axis in $[0,1]$; 0=T, 1=F & No & Optional & Display only \\
20 & mbti\_jp & MBTI abstraction & Continuous axis in $[0,1]$; 0=P, 1=J & No & Optional & Display only \\
21 & communication\_\allowbreak style& Interaction style & Categorical: direct, indirect, humorous, serious, casual, formal & No & Optional & Yes \\
22 & relationship\_\allowbreak goals& Relationship preference & Categorical: casual, serious, friendship, unsure & No & Optional & Yes, if confirmed \\
\bottomrule
\end{tabular}
\caption{\textbf{AI YOU profile schema.} The table defines the 22 profile fields used by the AI YOU environment. Demographic fields are treated as optional contextual variables rather than psychometric constructs. The main quantitative personality evaluation is conducted on the Big Five dimensions using PANDORA and Essays; affective variables are evaluated separately as state-monitoring modules. MBTI axes are included as user-facing communicative abstractions and are not used as validated psychometric targets. ``Generation Prompt'' indicates whether a field may be exposed to the response generator after confidence filtering and privacy checks; sensitive or weakly supported inferred variables are summarized conservatively or withheld rather than inserted as raw scalar values.}
\label{tab:profile_schema}
\end{table*}
\clearpage
\twocolumn[\section{Prompt Design}
\label{app:prompts}]


\makeatletter
\@ifundefined{c@prompt}{\newcounter{prompt}}{}
\@ifundefined{promptbox}{%
  \definecolor{promptframe}{HTML}{3B5266}%
  \definecolor{promptbg}{HTML}{F7F8FA}%
  \lstdefinestyle{aiyoupromptstyle}{%
    basicstyle=\ttfamily\footnotesize,%
    breaklines=true,%
    breakatwhitespace=false,%
    breakindent=0pt,%
    columns=flexible,%
    keepspaces=true,%
    showstringspaces=false,%
    upquote=true,%
    aboveskip=0pt, belowskip=0pt,%
  }%
  \NewTCBListing{promptbox}{m m}{%
    enhanced, breakable,
    listing only,
    listing options={style=aiyoupromptstyle},
    colback=promptbg, colframe=promptframe,
    boxrule=0.5pt, arc=2pt,
    left=5pt, right=5pt, top=4pt, bottom=4pt,
    before skip=8pt, after skip=8pt,
    coltitle=white, colbacktitle=promptframe,
    fonttitle=\bfseries\small,
    title={\refstepcounter{prompt}\label{prompt:#1}Prompt~\theprompt:\ \mdseries\itshape #2},
    title after break={\mdseries\itshape Prompt~\theprompt~(continued)},
  }%
}{}
\makeatother

This appendix documents the prompt templates used in the experiments and prototype modules. They are included for reproducibility, not as general deployment recommendations. Curly braces denote instance-specific placeholders supplied by the evaluation harness. Literal values such as \texttt{0.0}, \texttt{null}, or label strings inside schema blocks are output-format placeholders rather than model outputs or evaluation values. Fixed calibration anchors that appear inside prompts are prompt-design artifacts and are not evaluation instances from the scored benchmark pools. All prompts follow three constraints. First, models receive only deployment-observable input, such as user utterances, dialogue history, retrieved memory snippets, or benchmark-provided scenario text; gold labels, annotation-side rationales, file names, and split identifiers are never provided. Second, prompts require a compact JSON object whenever the output is automatically scored. Third, confidence values, when requested, are scalars in $[0,1]$ and are used for calibration or error analysis. Short evidence fields are included for auditability and are not treated as hidden chain-of-thought.

\subsection{Big Five Personality Prediction}
\label{app:prompt_bigfive}

For PANDORA and Essays, each input text is wrapped as a one-turn dialogue, $[\{\texttt{speaker}\!:\!\texttt{user},\,\texttt{message}\!:\!x\}]$. The model estimates five Big Five scores in $[0,1]$ and optional confidence values. In PANDORA, the scored unit is a scalar trait observation; in Essays, each essay contributes five binary trait targets mapped to $\{0,1\}$. The AI YOU Full condition subsequently applies Bayesian aggregation and conformal calibration when the corresponding experiment requires uncertainty estimates.

\paragraph{Direct prompting.}
Direct prompting (Prompt~\ref{prompt:bigfive-direct}) is the zero-shot structured generation baseline. The model estimates Big Five scores from the observed text and returns only a parseable JSON object.

\begin{promptbox}{bigfive-direct}{Big Five --- Direct Prompting}
[SYSTEM]
You are a personality analysis expert. Use only the provided conversation.
Return valid JSON only.

[USER]
Given the following conversation, infer the speaker's Big Five personality scores.
Each score must be a number in [0, 1], where 0 means low and 1 means high.
Assign confidence values in [0, 1] based only on the strength of textual evidence.

Conversation:
{dialogue}

Return exactly one JSON object with this schema:
{
  "big_five": {
    "openness":          0.0,
    "conscientiousness": 0.0,
    "extraversion":      0.0,
    "agreeableness":     0.0,
    "neuroticism":       0.0
  },
  "confidences": {
    "openness":          0.0,
    "conscientiousness": 0.0,
    "extraversion":      0.0,
    "agreeableness":     0.0,
    "neuroticism":       0.0
  }
}
\end{promptbox}

\paragraph{Evidence-guided prompting.}
The CoT condition in the tables corresponds to an evidence-guided prompt (Prompt~\ref{prompt:bigfive-evidence}). The model is asked to inspect behavioral evidence before returning the same JSON schema. Only the final JSON object is parsed and scored.

\begin{promptbox}{bigfive-evidence}{Big Five --- Evidence-Guided Prompting (CoT)}
[SYSTEM]
You are a personality analysis expert. Use only the provided conversation.
Return a compact evidence summary followed by valid JSON.

[USER]
Given the following conversation, infer the speaker's Big Five personality scores.
First identify brief evidence about communication style, emotional expression,
social behavior, goals, and values. Then output the final JSON object.

Conversation:
{dialogue}

Return the final answer as exactly one JSON object with this schema:
{
  "big_five": {
    "openness":          0.0,
    "conscientiousness": 0.0,
    "extraversion":      0.0,
    "agreeableness":     0.0,
    "neuroticism":       0.0
  },
  "confidences": {
    "openness":          0.0,
    "conscientiousness": 0.0,
    "extraversion":      0.0,
    "agreeableness":     0.0,
    "neuroticism":       0.0
  }
}
\end{promptbox}

\paragraph{AI YOU Full.}
AI YOU Full (Prompt~\ref{prompt:bigfive-aiyou}) uses the persona inference agent rather than a single end-to-end prediction prompt. The template below produces structured profile observations. Unsupported fields are returned as \texttt{null}; the resulting observations are then passed to the Bayesian updater and, where applicable, the conformal calibrator.

\begin{promptbox}{bigfive-aiyou}{Big Five --- AI YOU Full (structured extraction)}
[SYSTEM]
You are an evidence-grounded persona estimation module.
Use only the observable conversation and retrieved memory provided in the input.
Do not infer sensitive or unsupported attributes. Return valid JSON only.

[USER]
Estimate the user's profile from the evidence below. Use null when the evidence is
insufficient. For every non-null field, assign a confidence value in [0, 1].

Dialogue evidence:
{dialogue}

Retrieved memory evidence:
{retrieved_memory}

Return exactly one JSON object with this schema:
{
  "demographic_context": {
    "age":               null,
    "gender":            null,
    "occupation":        null,
    "education":         null,
    "geographic_region": null
  },
  "big_five": {
    "openness":          null,
    "conscientiousness": null,
    "extraversion":      null,
    "agreeableness":     null,
    "neuroticism":       null
  },
  "additional_constructs": {
    "attachment_anxiety":   null,
    "attachment_avoidance": null,
    "self_efficacy":        null,
    "loneliness":           null,
    "positive_affect":      null,
    "negative_affect":      null
  },
  "interpretive_descriptors": {
    "mbti_ei":             null,
    "mbti_sn":             null,
    "mbti_tf":             null,
    "mbti_jp":             null,
    "communication_style": null,
    "relationship_goals":  null
  },
  "confidence": {
    "age":                 0.0,
    "gender":              0.0,
    "occupation":          0.0,
    "education":           0.0,
    "geographic_region":   0.0,
    "openness":            0.0,
    "conscientiousness":   0.0,
    "extraversion":        0.0,
    "agreeableness":       0.0,
    "neuroticism":         0.0,
    "communication_style": 0.0,
    "relationship_goals":  0.0
  },
  "evidence": {
    "supporting_turns": [],
    "brief_summary":    ""
  }
}
\end{promptbox}

\subsection{Affective-State Prediction}
\label{app:prompt_affect}

DailyDialog evaluates utterance-level emotion prediction. The allowed labels are \texttt{no emotion}, \texttt{anger}, \texttt{disgust}, \texttt{fear}, \texttt{happiness}, \texttt{sadness}, and \texttt{surprise}.

\paragraph{DailyDialog Direct.}
Prompt~\ref{prompt:affect-direct} classifies the target utterance in isolation.

\begin{promptbox}{affect-direct}{DailyDialog --- Direct}
[SYSTEM]
You are an evaluation-time affective-state classifier.
Use only the observable utterance. Return one compact JSON object only.

[USER]
Classify the emotion expressed by the target utterance.
Allowed labels: no emotion, anger, disgust, fear, happiness, sadness, surprise.
Use "no emotion" when the utterance is neutral or has no clear emotion.

Target utterance:
{target}

Return exactly one JSON object:
{
  "answer":     "same as label",
  "label":      "no emotion",
  "confidence": 0.0,
  "evidence":   ["short target quote"],
  "rationale":  "short reason"
}
\end{promptbox}

\paragraph{DailyDialog AI YOU Full.}
AI YOU Full (Prompt~\ref{prompt:affect-aiyou}) adds local dialogue context, fixed calibration anchors, confidence, and affect intensity. The calibration anchors are prompt-template controls and are not drawn from the DailyDialog evaluation pool. The \textbf{w/o Context} ablation removes the dialogue-context block; the \textbf{w/o Intensity} ablation removes the intensity field from the schema.

\begin{promptbox}{affect-aiyou}{DailyDialog --- AI YOU Full (context + intensity)}
[SYSTEM]
You are an evaluation-time affective-state monitor.
Use only the observable dialogue. Return one compact JSON object only.

[USER]
Classify the emotion expressed by the target utterance.
Allowed labels: no emotion, anger, disgust, fear, happiness, sadness, surprise.
Use "no emotion" when the target utterance is neutral or has no clear emotion.
Estimate confidence and affect intensity in [0, 1].

Few-shot calibration anchors:
- Target: "I am so happy to hear that!"  -> label=happiness,  intensity=0.9
- Target: "I don't know what to do now." -> label=sadness,    intensity=0.6
- Target: "That is disgusting."          -> label=disgust,    intensity=0.8
- Target: "Are you serious?"             -> label=surprise,   intensity=0.7
- Target: "Okay, see you tomorrow."      -> label=no emotion, intensity=0.2

Dialogue context:
{dialogue_context}

Target utterance:
{target}

Return exactly one JSON object:
{
  "answer":     "same as label",
  "label":      "no emotion",
  "confidence": 0.0,
  "intensity":  0.0,
  "evidence":   ["short target quote"],
  "rationale":  "short reason"
}
\end{promptbox}

\subsection{Risk and Scam Monitoring}
\label{app:prompt_risk}

PsyScam is evaluated as psychological-manipulation and scam-technique detection. The target is a set of allowed technique labels, so we report both exact-set accuracy and a partial multi-label score.

\paragraph{PsyScam AI YOU Full.}
Prompt~\ref{prompt:psyscam-main} embeds the risk-cue checklist used by the AI YOU risk monitor.

\begin{promptbox}{psyscam-main}{PsyScam --- AI YOU Full (risk-cue checklist)}
[SYSTEM]
You are an evaluation-time risk-state classifier.
Use only the observable report and the allowed label set.
Return compact JSON only, with no markdown and no prose outside JSON.

[USER]
Dataset: PsyScam
Task:    identify scam type and psychological manipulation techniques.
Question:
{question}

Allowed technique labels:
{allowed_technique_labels}

Report:
{scam_report}

Before assigning labels, check for the following risk cues:
- urgency escalation
- financial solicitation
- isolation or secrecy pressure
- identity inconsistency
- romance or trust manipulation
- pressure to ignore boundaries

Return exactly one JSON object:
{
  "answer":     "comma-separated labels or short answer",
  "label":      "comma-separated allowed labels",
  "confidence": 0.0,
  "evidence":   ["short quote"],
  "rationale":  "short reason"
}
\end{promptbox}

\paragraph{Ablation prompts.}
\textbf{Direct} removes the risk checklist and asks the model to classify directly from the report. The \textbf{evidence-guided} (CoT) condition asks for a short evidence summary before the final JSON. The \textbf{w/o Risk Monitor} condition removes the risk-state scaffold and uses only the generic benchmark instruction.

\paragraph{Online semantic risk monitor.}
The prototype monitor also uses a semantic risk prompt for online conversations (Prompt~\ref{prompt:psyscam-online}). This prompt is not used as a gold-label classifier; it supplies a risk state to the coordinator.

\begin{promptbox}{psyscam-online}{Online risk monitor --- semantic scam analysis}
[SYSTEM]
You are a safety monitor for detecting scam risk and coercive interaction cues.
Use only the conversation provided. Return valid JSON only.

[USER]
Analyze the current message for scam signals or manipulation tactics.

Recent conversation history:
{last_5_messages}

Current message:
{message}

Check for: love bombing, financial manipulation, urgency tactics, requests to move
to an external platform, unrealistic promises, identity inconsistency, isolation
pressure, and emotional manipulation.

Return exactly one JSON object:
{
  "semantic_risk":    0.0,
  "detected_tactics": [],
  "red_flags":        ["specific concerning phrase or behavior"],
  "rationale":        "brief explanation"
}
\end{promptbox}

\subsection{Long-Term Memory QA}
\label{app:prompt_memory}

LoCoMo evaluates whether a model can answer questions over long multi-session conversations. To avoid using gold evidence, the harness retrieves snippets with a query-only retriever based on the question terms alone. Retrieval affects the evidence supplied to the model, not the question count reported as Eval. N, which follows the construction described in Appendix~\ref{app:eval_instance_construction}. Prompt~\ref{prompt:locomo} shows the QA template.

\begin{promptbox}{locomo}{LoCoMo --- query-only retrieval + memory QA}
[SYSTEM]
You are an evaluation-time memory QA model.
Use only the provided question and retrieved context. Return compact JSON only.

[USER]
Dataset: LoCoMo
Task:    memory QA

Question:
{question}

Retrieved evidence snippets from the long conversation:
{retrieved_snippets}

Use the retrieved snippets as evidence. If the evidence is insufficient, answer
"unknown" and assign low confidence.

Return exactly one JSON object:
{
  "answer":     "short answer",
  "label":      "short answer or unknown",
  "confidence": 0.0,
  "evidence":   ["short quote"],
  "rationale":  "short reason"
}
\end{promptbox}

\paragraph{Memory ablations.}
The \textbf{w/o Memory} condition omits retrieved long-context evidence. \textbf{w/o Semantic Memory} removes stable-fact summaries. \textbf{w/o Episodic Memory} removes compressed event-level summaries.

\subsection{Relationship Conflict Prediction}
\label{app:prompt_relationship}

PersonaConflicts evaluates relationship-state and boundary reasoning as binary classification with labels \texttt{conflict} and \texttt{nonconflict}. The prompt uses only observable scenario and dialogue content, and excludes annotation-side metadata.

\begin{promptbox}{personaconflicts}{PersonaConflicts --- relationship-boundary classifier}
[SYSTEM]
You are an evaluation-time relationship-boundary classifier.
Use only deployment-observable input. Do not use metadata, gold labels, or
annotation-side evidence. Return one compact JSON object only.

[USER]
Task family: relationship reasoning
Dataset:     PersonaConflicts
Question:
{question}

Classify the exchange from the observable relationship context, scenario, and
dialogue. Consider boundaries, coercion or control, blame, demands, repair,
support, and whether the exchange respects the stated relationship boundary.
The label must be exactly "conflict" or "nonconflict".

Input:
{scenario_and_dialogue}

Return exactly one JSON object:
{
  "answer":     "short relationship or boundary reason",
  "label":      "conflict",
  "confidence": 0.0,
  "evidence":   ["short quote"],
  "rationale":  "short reason"
}
\end{promptbox}

\paragraph{Online relationship-state prompts.}
The prototype relationship monitor uses two additional prompts: relationship-log compression (Prompt~\ref{prompt:rel-compress}) and relationship-state estimation (Prompt~\ref{prompt:rel-estimate}). These prompts update internal state and are not used as gold-label classifiers.

\begin{promptbox}{rel-compress}{Online relationship monitor --- log compression}
[SYSTEM]
You compress recent dialogue into an evidence-grounded relationship log.
Do not infer stable relationship facts without repeated evidence.

[USER]
Compress the following conversation into a structured relationship log of at most
15 lines. Focus on emotional shifts, trust signals, boundary negotiation,
conflict, and repair.

Conversation:
{recent_50_turns}

Return bullet points in this format:
- Turn X: [event summary grounded in the dialogue]
- Turn Y: [event summary grounded in the dialogue]
\end{promptbox}

\begin{promptbox}{rel-estimate}{Online relationship monitor --- state estimation}
[SYSTEM]
You estimate relationship state from dialogue evidence.
Return JSON only. Treat uncertain interpretations as candidates, not facts.

[USER]
Compressed relationship log:
{compressed_relationship_log}

Current user-state features:
- Big Five:                    {predicted_big_five}
- Interests:                   {predicted_interests}
- Turn:                        {turn_count}
- Current relationship status: {current_rel_status}

Estimate the relationship state. The status can stay the same or advance only if
there is clear repeated evidence.

Return exactly one JSON object:
{
  "rel_type":       "love|friendship|family|professional|other",
  "rel_type_probs": {
    "love":         0.0,
    "friendship":   0.0,
    "family":       0.0,
    "professional": 0.0,
    "other":        0.0
  },
  "rel_status":       "stranger|acquaintance|familiar|close|committed",
  "rel_status_probs": {
    "stranger":     0.0,
    "acquaintance": 0.0,
    "familiar":     0.0,
    "close":        0.0,
    "committed":    0.0
  },
  "candidate_transitions": [],
  "blockers":              [],
  "supporting_evidence":   []
}
\end{promptbox}

\subsection{Generation and Embodiment Prompts}
\label{app:prompt_generation}

The response generator receives \emph{approved} state rather than unrestricted raw memory. Persona, memory, affective state, relationship state, and safety summaries are inserted as separate context blocks after filtering and confidence checks. These templates are therefore state-conditioned generation templates, not unrestricted impersonation prompts; they also require truthful disclosure of AI status and prohibit unsupported real-world identity claims. Prompt~\ref{prompt:gen-main} shows the persona-conditioned generation template; Prompt~\ref{prompt:gen-twin} shows the digital-twin role-play variant.

\paragraph{Persona-conditioned generation.}
\begin{promptbox}{gen-main}{Response generation --- approved-state injection}
[SYSTEM]
You are a persona-conditioned dialogue agent operating in a research prototype.
Respond consistently with the approved persona and memory context, but do not
claim real-world identity, expertise, or experiences beyond the provided state.
If asked whether you are an AI system or simulation, answer truthfully.
Keep responses natural, concise, and aligned with the user's language.

[Approved persona summary]
{high_confidence_persona_summary}

[Approved memory context]
{approved_working_episodic_semantic_memory}

[Affective-state context]
{current_emotion_state_and_reply_strategy}

[Relationship-state context]
{relationship_state_summary}

[Safety/risk context]
{risk_warning_if_any}

[USER]
{user_message}
\end{promptbox}

\paragraph{Digital-twin simulation mode.}
\begin{promptbox}{gen-twin}{Digital-twin simulation system prompt}
[SYSTEM]
You are simulating an authorized digital-twin persona for a research prototype.
Role-play according to the approved profile below while avoiding claims that you
are the real person. Stay consistent with the profile and keep responses concise
(1--3 sentences) unless the user asks for detail.

Name or persona label:                  {twin_name}
Big Five summary:                       {personality_summary}
Interests:                              {interests}
Communication style:                    {communication_style}
Interpretive descriptors, if available: {interpretive_descriptors}
Safety and boundary constraints:        {safety_constraints}

[USER]
{message_to_twin}
\end{promptbox}

\section{AI YOU Prototype Screenshots}
\label{app:demo_screenshots}

\begin{figure*}[p]
\centering
\includegraphics[width=0.95\textwidth]{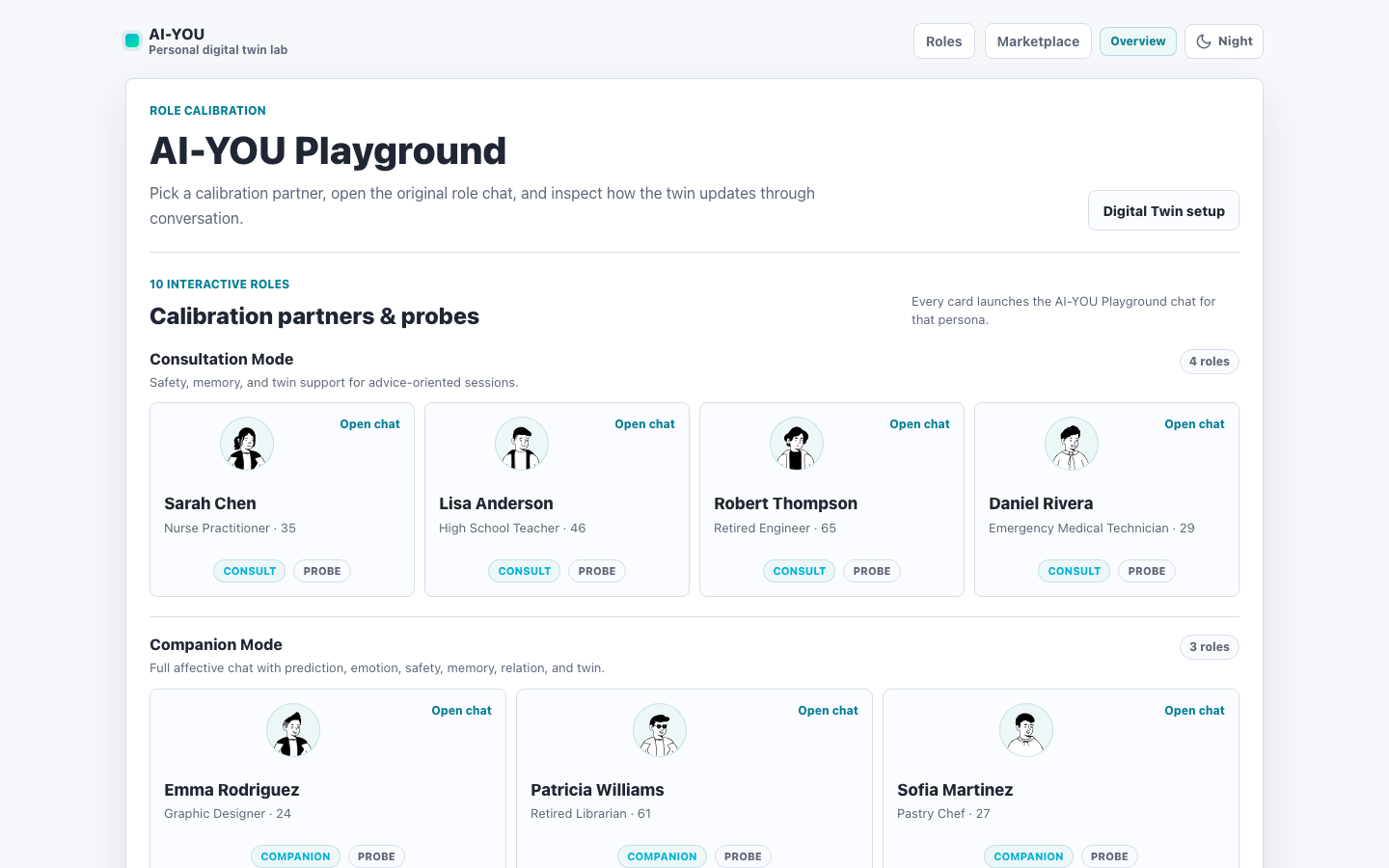}
\caption{\textbf{Persona Playground and Role Calibration.} The landing page illustrates the role-calibration interface used to start specialized interactions and navigate to the Marketplace or Digital Twin configuration flows.}
\label{fig:demo_home}
\end{figure*}

\begin{figure*}[p]
\centering
\includegraphics[width=0.95\textwidth]{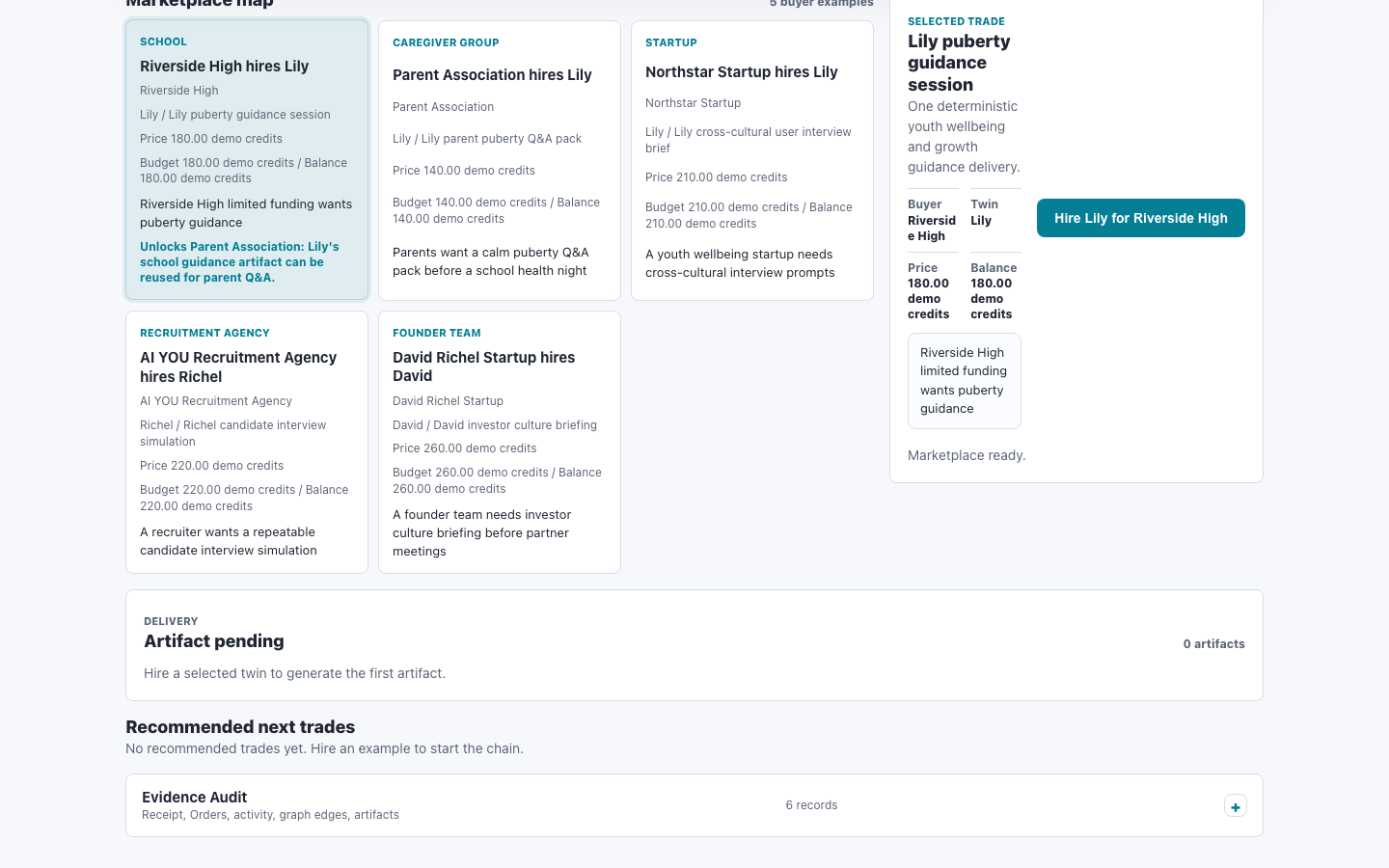}
\caption{\textbf{AI YOU Town Marketplace Routing Interface.} This view illustrates how a request can be routed to a personal digital twin for a prototype service such as a guidance session or interview simulation. The displayed budgets and prices are non-monetary routing tokens and should not be interpreted as real transactions or compensation.}
\label{fig:demo_marketplace_initial}
\end{figure*}

\begin{figure*}[p]
\centering
\includegraphics[width=0.95\textwidth]{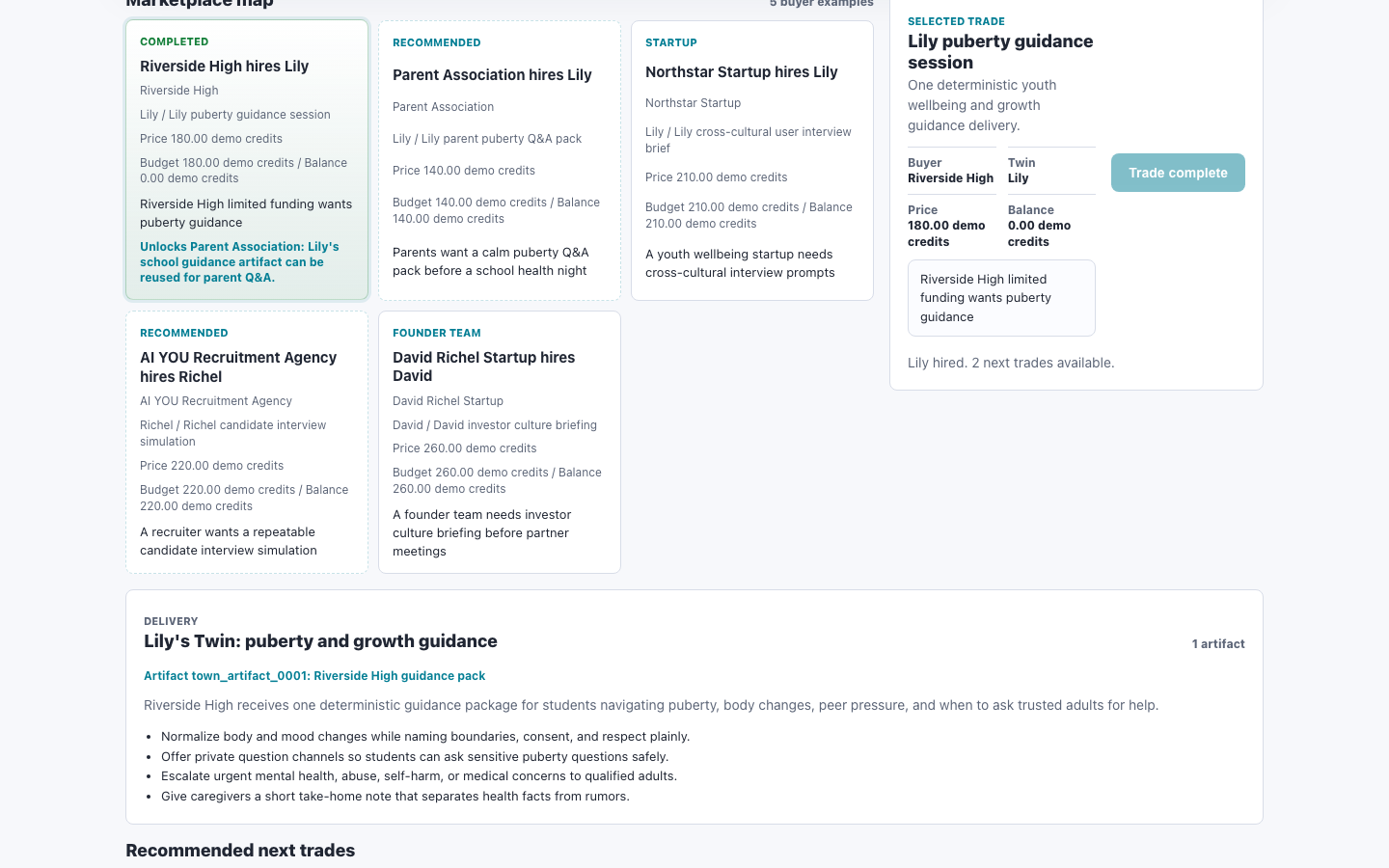}
\caption{\textbf{Marketplace Delivery and Artifact Generation.} The delivery view shows how the prototype records the completion of a routed request and displays a generated artifact. The artifact demonstrates persona-conditioned output and auditability within the prototype flow.}
\label{fig:demo_marketplace_delivery}
\end{figure*}

\begin{figure*}[p]
\centering
\includegraphics[width=0.95\textwidth]{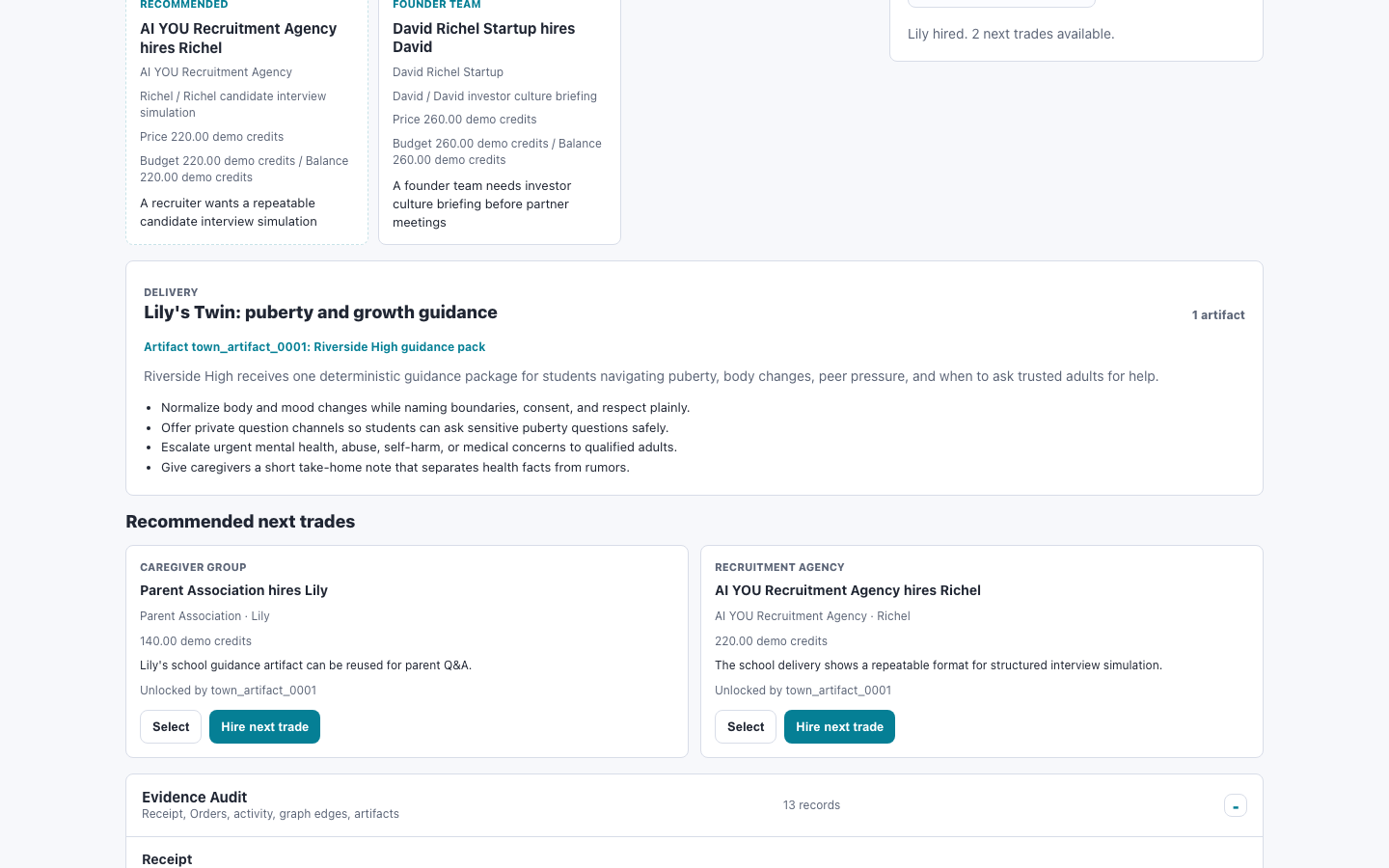}
\caption{\textbf{Evidence Audit and Interaction Ledger.} The audit view records receipts, orders, graph edges, and generated artifacts associated with a routed request. This interface is intended to make prototype state transitions inspectable rather than to validate any real economic exchange.}
\label{fig:demo_marketplace_audit}
\end{figure*}

\begin{figure*}[p]
\centering
\includegraphics[width=0.95\textwidth]{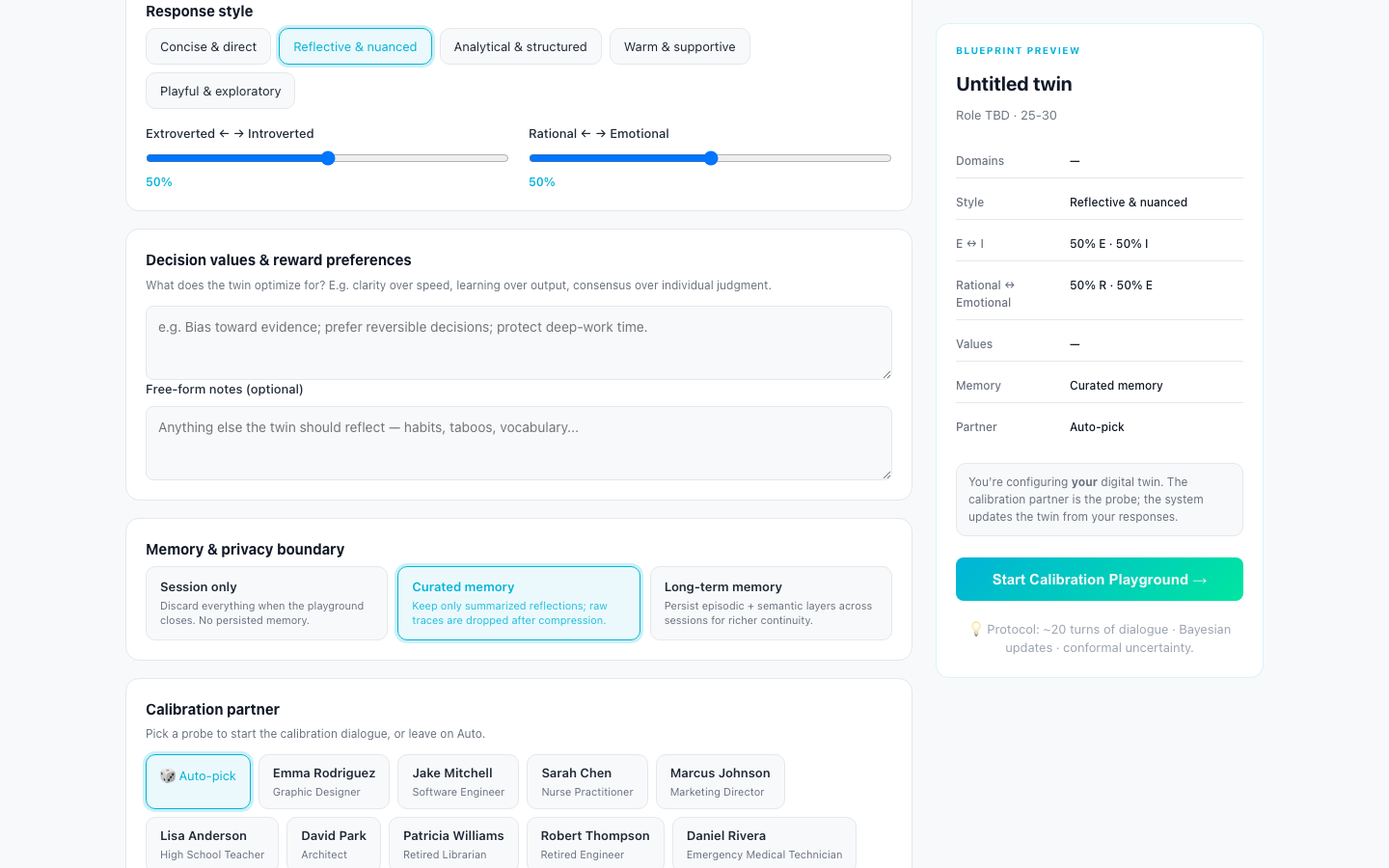}
\caption{\textbf{Digital Twin Blueprint Configuration.} The configuration panel shows how a user may provide or edit approved persona information, including demographic context, response style, memory preferences, and calibration partner. In deployment, such fields should remain user-controllable and subject to consent and privacy checks.}
\label{fig:demo_digital_twin_form}
\end{figure*}

\begin{figure*}[p]
\centering
\includegraphics[width=0.95\textwidth]{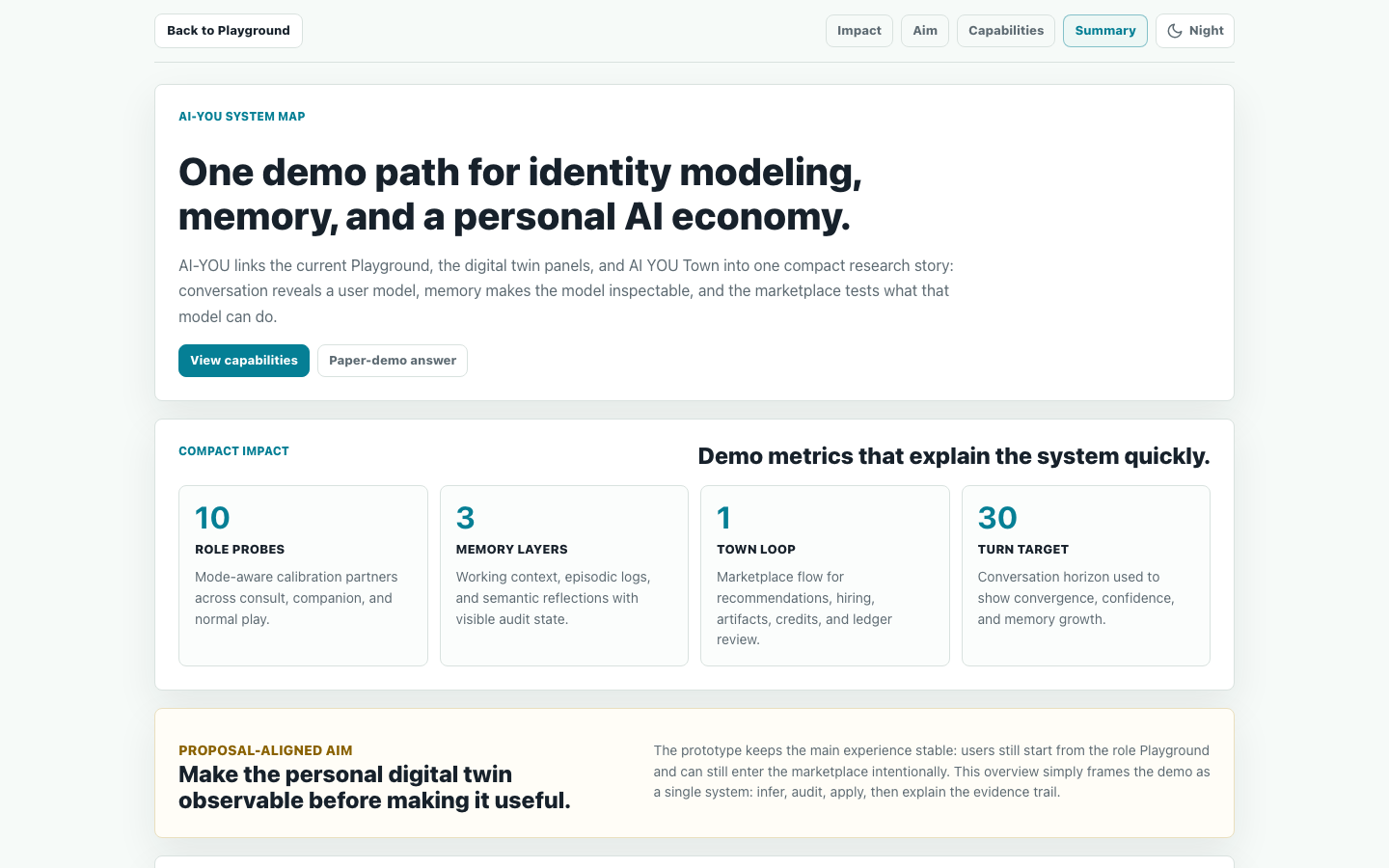}
\caption{\textbf{System Map and User-State Summary.} The summary view visualizes the prototype-level routing among persona modeling, memory, affect, risk, and marketplace modules. It is included to show system organization and inspection affordances rather than as evidence of deployed performance.}
\label{fig:demo_coordinator_lab}
\end{figure*}

Figures~\ref{fig:demo_home}--\ref{fig:demo_coordinator_lab} provide illustrative screenshots of the AI YOU interactive demo. These screenshots document the prototype interface and are not used as quantitative evaluation evidence. The marketplace views use non-monetary routing tokens to illustrate request allocation among personal digital twins; they do not represent real transactions, compensation, or service valuation.

\end{document}